\documentclass[9pt,journal,cspaper,compsoc]{IEEEtran}

\usepackage{times}
\usepackage{epsfig}
\usepackage{graphicx}
\usepackage{subcaption}
\usepackage{amsmath}
\usepackage{amssymb}
\usepackage{color}
\usepackage{tabularx}
\usepackage{multirow}
\usepackage[abs]{overpic}
\usepackage{algorithm}
\usepackage{algorithmic}
\usepackage{wrapfig}
\usepackage{ragged2e}
\usepackage[nocompress]{cite}
\usepackage[colorlinks,bookmarksnumbered,bookmarksopen]{hyperref}

\usepackage{color}

\definecolor{R}{RGB}{202,0,0} 
\definecolor{B}{RGB}{0,102,204} 

\usepackage{xcolor}
\usepackage{colortbl}

\usepackage{makecell}
\usepackage{booktabs,multirow,rotating}

\usepackage{colortbl}

\usepackage{ragged2e}

\usepackage{amsmath}
\usepackage{amssymb}
\usepackage{diagbox}

\usepackage{bbding}
\usepackage{pifont}
\usepackage{booktabs}
\usepackage[colorlinks,bookmarksnumbered,bookmarksopen]{hyperref}

\usepackage{geometry}
\graphicspath{{./Imgs/}{./CVPR/Imgs/}}
\DeclareGraphicsExtensions{.pdf,.png,.jpg}
\newcolumntype{C}[1]{>{\centering\arraybackslash}p{#1}}

\begin{document}
\title{Colorectal Polyp Segmentation in the \\ Deep Learning Era: A Comprehensive Survey}

\author{Zhenyu Wu, Fengmao Lv, Chenglizhao Chen, Aimin Hao, Shuo Li 
\IEEEcompsocitemizethanks{
\IEEEcompsocthanksitem Zhenyu Wu and Fengmao Lv are with
the Southwest Jiaotong University.

\IEEEcompsocthanksitem  Chenglizhao Chen is with China University of Petroleum (Qingdao).
\IEEEcompsocthanksitem  Aimin Hao is with Beihang University
\IEEEcompsocthanksitem  Shuo Li is with Case Western Reserve University
}
}

\markboth{IEEE TRANSACTIONS ON PATTERN ANALYSIS AND MACHINE INTELLIGENCE}%
{Liu \MakeLowercase{\textit{et al.}}: Richer Convolutional Features for Edge Detection}

\IEEEtitleabstractindextext{%
\begin{abstract}
\justifying
Colorectal polyp segmentation (CPS), an essential problem in medical image analysis, has garnered growing research attention. Recently, the deep learning-based model completely overwhelmed traditional methods in the field of CPS, and more and more deep CPS methods have emerged, bringing the CPS into the deep learning era.
To help the researchers quickly grasp the main techniques, datasets, evaluation metrics, challenges, and trending of deep CPS, this paper presents a systematic and comprehensive review of deep-learning-based CPS methods from 2014 to 2023, a total of 115 technical papers. In particular, we first provide a comprehensive review of the current deep CPS with a novel taxonomy, including network architectures, level of supervision, and learning paradigm. More specifically, network architectures include eight subcategories, the level of supervision comprises six subcategories, and the learning paradigm encompasses 12 subcategories, totaling 26 subcategories. Then, we provided a comprehensive analysis the characteristics of each dataset, including the number of datasets, annotation types, image resolution, polyp size, contrast values, and polyp location. Following that, we summarized CPS's commonly used evaluation metrics and conducted a detailed analysis of 40 deep SOTA models, including out-of-distribution generalization and attribute-based performance analysis. Finally, we discussed deep learning-based CPS methods' main challenges and opportunities.

\end{abstract}

\begin{IEEEkeywords}
Colorectal Polyp Segmentation, Deep Learning, Colorectal Cancer.
\end{IEEEkeywords}}

\maketitle

\IEEEdisplaynontitleabstractindextext
\IEEEpeerreviewmaketitle

\begin{figure}[t]
\centering
\resizebox{0.48\textwidth}{!}{
\includegraphics[width=\textwidth]{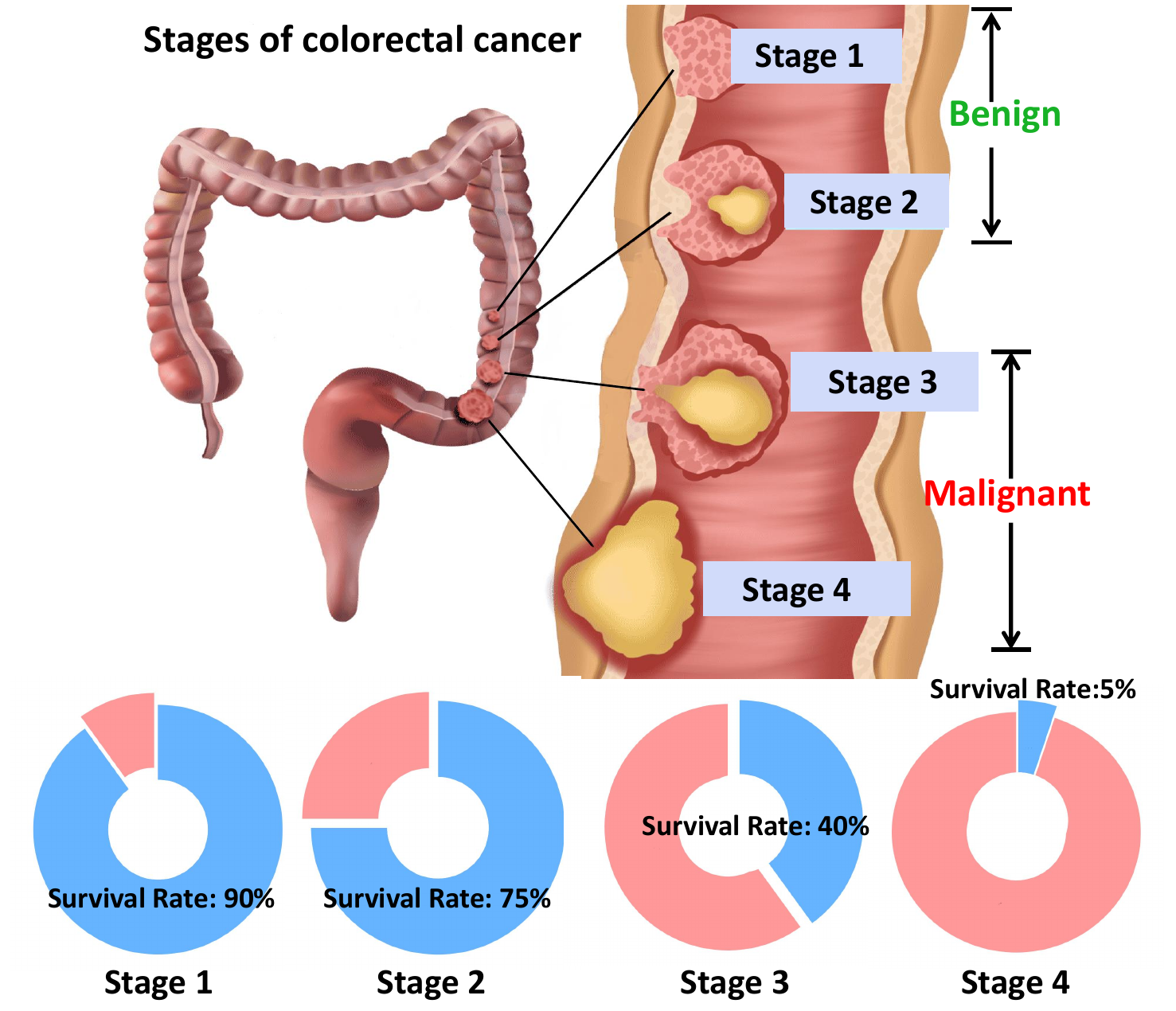}
}
\caption{Colorectal cancers originate from polyps, undergoing a progression from normal mucosa to adenomatous polyps, then pre-cancerous polyps, adenocarcinoma, and finally transforming into cancer. The survival rate of CRC patients exceeds 90\% in the first stage but drastically drops to less than 5\% in the last stage.} 
\label{fig:demo}
\end{figure}

\IEEEraisesectionheading{\section{Introduction}\label{sec:introduction}}
\IEEEPARstart{C}olorectal cancer (CRC) has become the second leading cause of cancer-related death worldwide, causing millions of incidence cases and deaths every year.
 Research shows that nearly 95\% of colorectal cancers originate from polyps, undergoing a progression from normal mucosa to adenomatous polyps, then precancerous polyps, adenocarcinoma, and finally transforming into cancer \cite{bernal2012towards}, as shown in Fig. \ref{fig:demo}.  The survival rate of CRC patients exceeds 90\% in the initial stage but drastically drops to less than 5\% in the last stage \cite{ji2022video}. Thus, the survival rate can be improved by early diagnosis and treatment of precancerous polyps via colonoscopy screening, such as colonoscopy and capsule endoscopy. 
Although colonoscopy provides information on the appearance and localization of polyps, it demands expensive labor resources and exhibits a high miss rate due to the following challenges: 1) Diagnostic accuracy heavily relies on the doctor's experience.2) Polyps exhibit significant variations in color, size, and shape. 3) The boundaries between polyps and surrounding tissues are unclear. For instance, sessile polyps, also known as flat polyps, is no significant elevation compared to the surrounding normal tissue, making them inconspicuous and challenging to detect. Therefore, it is essential to develop an automated and accurate colorectal polyp segmentation (CPS) system, minimizing the occurrence of misdiagnoses to the greatest extent.

Traditional polyp segmentation methods primarily focus on learning low-level features, such as texture, shape, or color distribution, failing to deal with complex
scenarios. With advanced deep learning, plenty of CNN/Transformer-based methods have been proposed for polyp segmentation. In recent years, UNet \cite{ronneberger2015u} based deep learning methods such as UNet++ \cite{zhou2018unet}, ResUNet++ \cite{jha2019resunet}, and PraNet \cite{fan2020pranet} have dominated the field. Recently, transformer-based models \cite{zhang2021transfuse,wang2022stepwise,ling2023probabilistic} have also been proposed for polyp segmentation and achieve state-of-the-art (SOTA) performance. Despite significant progress made by these deep learning models, it still lacks a comprehensive overview for the polyp segmentation task to date.

\begin{figure*}[t]
\centering
\resizebox{\textwidth}{!}{
\includegraphics[width=\textwidth]{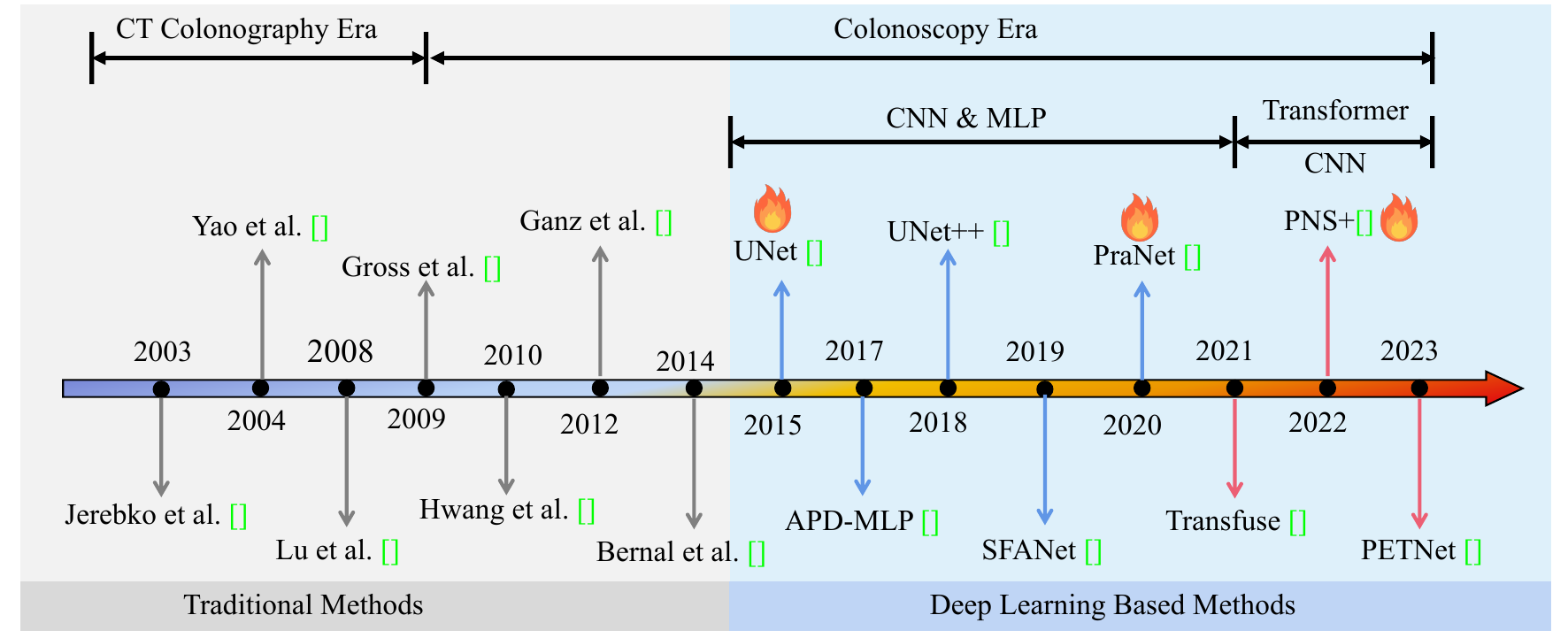}
}
\caption{A brief history of CPS. The first seminal work \cite{jerebko2003polyp} on CPS can be traced back to  2003. The first deep learning-based CPS method \cite{ronneberger2015u} emerged in 2015. The methods mentioned above serve as milestones, which are typically highly cited. See Section \ref{sec: history} for more details.}
\label{fig:history}
\end{figure*}

This paper presents a systematic and comprehensive review of deep-learning-based CPS methods. We first provide a comprehensive review of the current deep CPS with a novel taxonomy, including network architectures, level of supervision, and their learning paradigm. Then, we also provide a comprehensive analysis of the characteristics of each dataset, including the number of datasets, annotation types, image resolution, polyp size, contrast values, and polyp location. We conducted a comprehensive summary and analysis of 40 deep SOTA models, building an open and standardized evaluation benchmarking. Finally, we discussed the main challenges and opportunities of deep learning-based CPS methods.
We hope that this survey can assist researchers in quickly grasping the development history of polyp segmentation and attracting more researchers to join the field.

\subsection{History and Scope}

\subsubsection{A brife history of polyp segmentation}
\label{sec: history}

As shown in Fig. \ref{fig:history}, we briefly outline the history of colorectal polyp segmentation.
To our best knowledge, the first work of CPS can be traced back to the seminal works of \cite{jerebko2003polyp} published in 2003, which employs the canny edge detector to segment images of polyp candidates for computed tomography (CT) colonography. Subsequently, Yao \textit{et al.} \cite{yao2004colonic} proposed an automatic method to segment colonic polyps in CT colonography, which is based on a combination of knowledge-guided intensity adjustment and fuzzy c-mean clustering. Gross \textit{et al.} \cite{gross2009polyp} presented the first automatic polyp segmentation algorithm for colonoscopic narrow-band images. Hwang \textit{et al.} \cite{hwang2010polyp} presented an unsupervised method for the detection of polyps in wireless capsule endoscopy videos, which adopts watershed segmentation with a novel initial marker selection method based on Gabor texture features and K-means clustering. These non-deep learning-based polyp segmentation methods \cite{lu2008accurate,ganz2012automatic,bernal2014polyp} typically depend on manually extracted low-level features and traditional segmentation algorithms. However, manually extracting low-level features for polyp segmentation is inadequate for handling complex scenarios.

With deep learning technology completely overwhelms traditional methods in the field of computer vision, more and more deep CPS methods \cite{yuan2017automatic,zhou2018unet,ji2022video} have emerged since 2015, bringing the polyp segmentation task into the deep learning era. Ronneberger \textit{et al.} \cite{ronneberger2015u} proposed a UNet shape network for biomedical image segmentation and won the ISBI cell tracking challenge 2015 by a large margin. 
Later, Fang \textit{et al.} \cite{fang2019selective} proposed a selective feature aggregation network with the area and boundary constraints for CPS. 
Fan \textit{et al.} \cite{fan2020pranet} propose a parallel reverse attention network for accurate polyp segmentation in colonoscopy images.
Zhang \textit{et al.} \cite{zhang2021transfuse} propose a transformer-based parallel-in-branch architecture, which combines Transformers and CNNs in a parallel style. 
Recently, Ling \textit{et al.} \cite{ling2023probabilistic}  proposed a Gaussian-Probabilistic guided semantic fusion method that progressively fuses the probability information of polyp positions with the decoder supervised by binary masks.

\subsubsection{Scope of this survey.}
Clinical colorectal polyp screening can be divided at the task level into three classical tasks, i.e., polyp classification, polyp detection, and polyp segmentation. This work focuses on the polyp segmentation task and aims to systematically summarize deep polyp segmentation methods, commonly used polyp segmentation datasets, evaluation metrics, and model performance.
Although the development of the deep polyp segmentation model is only eight years, it has spawned hundreds of papers, making it impractical to review all of them. In this survey, we mainly focus on these influential papers published in mainstream journals and conferences of medical image analysis from 2019 to 2023. Besides, for completeness and better readability, some early relevant works ranging from 2014 to 2018 have also been mentioned briefly. Due to space limitations and the extent of our knowledge, we apologize to those authors whose works could not be included in this paper.

\subsection{Related Surveys}
\label{sec:related_work}

Table \ref{tab:related_reivews} lists existing surveys related to this work. Among them, Prasath \textit{et al}. \cite{prasath2016polyp} reviewed ``capsule endoscopy'' colorectal polyp detection and segmentation methods based on handcrafted features and discussed the existing challenges. Later, Taha \textit{et al}. \cite{taha2017automatic} reviewed endoscopy colorectal polyp detection models, which divide these models into shape, texture, and fusion features. Sanchez-Peralta \textit{et al}. \cite{sanchez2020deep} mainly focuses on the deep learning-based colorectal polyp detection, localization, and segmentation approaches. Besides,  they also summarize six widely used polyp detection datasets for training or benchmarking and 19 commonly used metrics for model evaluation. Finally, a more recently published survey \cite{elkarazle2023detection} summarized the commonly adopted network architectures, introduced the benchmark datasets and evaluation metrics, and discussed the current challenges. Although existing reviews provide insights from different perspectives, they also have some
shortcomings: \textbf{1)} \textit{Insufficient in the number of literature.} In the above-mentioned surveys, the most comprehensive survey \cite{prasath2016polyp} contains only 37 papers. \textbf{2)} \textit{Limited in time span.} For example, surveys \cite{prasath2016polyp} and \cite{taha2017automatic} focuse on polyp segmentation methods before 2016 while \cite{sanchez2020deep} review polyp segmentation methods from 2015 to 2018. \cite{elkarazle2023detection} reviews 20 papers, but 18 out of 20 were published in 2022. \textbf{3)} Lacking comprehensive evaluation and analysis for existing models and datasets. 


\begin{table*}[!t]
\renewcommand\arraystretch{1.2}
  \centering
  \resizebox{\textwidth}{!}{
  \begin{tabular} {p{180pt}|p{20pt}<{\centering}|p{230pt}}\toprule   
   Title& Year   &  Description \\ \midrule

   {Polyp detection and segmentation from video capsule endoscopy: A review \cite{prasath2016polyp}} & 2016             &  {This paper reviews traditional based capsule endoscopy polyps detection and segmentation methods before 2016 with \textbf{only 37 papers}.}     \\ \midrule

   {Automatic polyp detection in endoscopy videos: A survey \cite{taha2017automatic}} & 2017              &  {This paper reviews handcrafted features based polyps detection methods  before 2016 with \textbf{only 28 papers}.}     \\ \midrule
 
   {Deep learning to find colorectal polyps in colonoscopy:A systematic literature review \cite{sanchez2020deep}} & 2020     &  {This paper reviews deep learning-based polyps detection, localization, and segmentation methods from 2015 to 2018 with  \textbf{only 35 papers}.}     \\  \midrule

   {Detection of colorectal polyps from colonoscopy using machine learning: A survey on modern techniques \cite{elkarazle2023detection}} & 2023    &  {This paper reviews deep learning-based polyps detection with \textbf{only 20 papers}, of which 18 out of 20 were published in 2022.}     \\  \midrule

  \end{tabular}}
  \vspace{-0.5em}
  \caption{Summary of previous surveys on CPS. For each survey, the title, publication year, and a summary of the content are provided. More discussion can be found in Section \ref{sec:related_work}. }
  \label{tab:related_reivews}
\end{table*}

\begin{figure*}[t]
\centering
\resizebox{\textwidth}{!}{
\includegraphics[width=\textwidth]{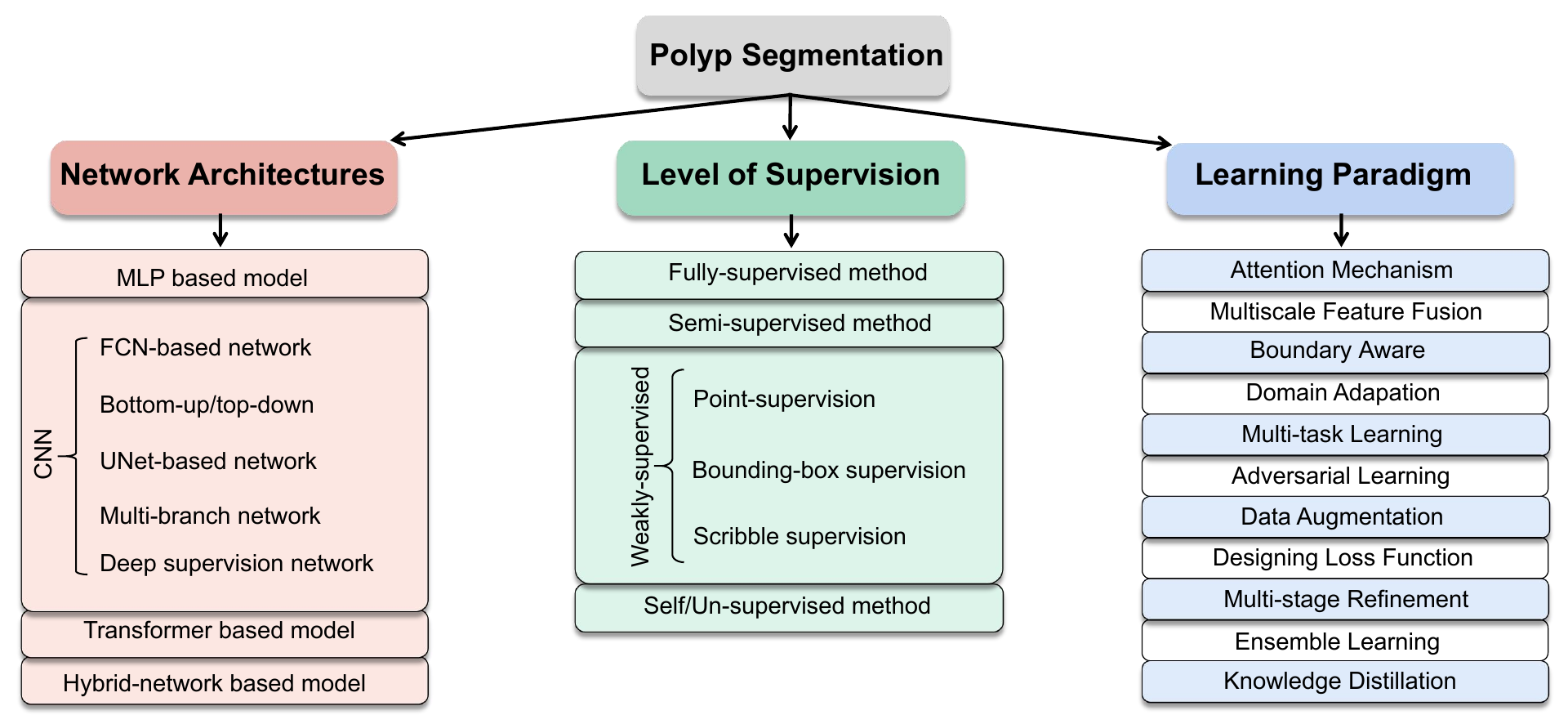}
}

\caption{An overview of our proposed taxonomy. We categorize existing polyp segmentation approaches into three branches from different perspectives: network architectures, level of supervision, and learning paradigm. Specifically, network architectures include eight subcategories, the level of supervision comprises six subcategories, and the learning paradigm encompasses 12 subcategories, totaling 26 subcategories.} 
\label{fig:taxonomy}
\end{figure*}

Recently, Mei \textit{et al.} \cite{mei2023survey} also conducted a polyp segmentation survey and released their paper on the Arxiv platform. Compared to this contemporary work and previous surveys, our work has several advantages:
\noindent  \textbf{1) A well-organized taxonomy of deep CPS models.} As shown in Fig. \ref{fig:taxonomy}, our survey categorized polyp segmentation methods into three main groups: network architectures, level of supervision, and learning paradigm. More specifically, network architectures include eight subcategories, the level of supervision comprises six subcategories, and the learning paradigm encompasses 12 subcategories, totaling 26 subcategories. In contrast, Mei's survey divided existing polyp segmentation methods into six categories.
\noindent  \textbf{2) A systematic study of current deep CPS methods.} As shown in Table \ref{tab:methods1} and Table \ref{tab:methods2}, our survey contains 115 papers from 2014 to 2023, while Mei's survey only covered 45 papers from 2019 to 2023. The number of papers in our survey is nearly three times that of theirs. Besides we also provided a summary of the architecture, key technologies, and training datasets utilized by the models proposed in each paper, aiming to enhance readers' understanding. 
\noindent  \textbf{3) A comprehensive analysis of current CPS datasets.} As shown in Fig. \ref{fig:heatmap}, Fig. \ref{fig:contrast_size} and Table \ref{tab:datasets}, we provide a comprehensive analysis of the characteristics of each dataset, including the number of datasets, annotations types, image resolution, polyp size, contrast values, number of polyps and polyp location. \textbf{4) An extensive performance comparison and analysis of deep CPS approaches.} As shown in Table \ref{tab:results}, our survey comprehensively summarized and analyzed 40 models from 2015 to 2023, while Mei's survey included only 21 models. In addition, as shown in Table \ref{tab:generalization} and Table \ref{tab:attribution-analysis}, We also assessed the model's generalization performance on the out-of-distribution dataset (i.e., PolypGen \cite{ali2023multi}) and its attribute-based performance on SUN-SEG\cite{ji2022video} datasets for better uncovering the nuanced strengths and weaknesses of deep CPS
models. \textbf{5) A deeper look into the challenges of CPS and providing some promising directions.} We extensively summarized the challenges encountered in polyp segmentation, including the deep model's interpretability, generalization ability, robustness to adversarial attacks, data privacy, domain shift, etc. We also highlighted several promising directions for CPS, such as combining CPS with unsupervised anomaly localization, large visual models and large language models.

\subsection{Our Contributions}
In summary, this survey systematically studies deep polyp segmentation methods, datasets, and metrics from 2014 to 2023, including 115 technical papers, 14 polyp segmentation datasets, and 12 commonly used evaluation metrics. We aim to provide an intuitive understanding and high-level insights into polyp segmentation so that researchers can choose the proper directions to explore. The contributions of this work are summarized as follows:

\begin{itemize} 
  \item \textbf{A systematic study of deep CPS methods from 2014 to 2023, a total of 115 technical papers.} We propose a novel taxonomy of polyp segmentation approaches and categorize these approaches into three main categories: network architectures, level of supervision, and learning paradigm, which can be further divided into 26 subcategories. The introduced taxonomy was designed to help researchers gain a deeper insight into the essential characteristics of deep CPS models.
  \item  \textbf{A comprehensive analysis on current CPS datasets.} We thoroughly analyze each CPS dataset, including the number of datasets, annotation types, image resolution, polyp size, contrast values, number of polyps, and polyp location. Furthermore, we analyzed the challenges and difficulties in the existing CPS dataset.
  \item \textbf{An extensive performance comparison and analysis of deep CPS models.} We conducted a comprehensive summary and analysis of 40 models from 2015 to 2023 for building an open and standardized evaluation benchmarking. Additionally, we evaluated the model's generalization performance on the multi-center dataset (PolypGen) and its attribute-based performance on SUN-SEG datasets to better understand the strengths and weaknesses of deep CPS models.

  \item \textbf{An in-depth delving into the challenges of deep CPS models and providing promising directions.} We summarized the challenges of current deep CPS models, including deep model's interpretability, generalization ability, robustness to adversarial attacks, data privacy, domain shift, etc. We also present several promising directions for CPS, such as combining CPS with unsupervised anomaly localization, large visual models, and large language models.
\end{itemize}

The rest of this survey is organized as follows: Section \ref{sec:method} presents a novel taxonomy and reviews some representative state-of-the-art deep learning-based polyp segmentation models for each category. Section \ref{sec:datasets} overviews some of the most popular polyp segmentation datasets and their characteristics. Section \ref{sec:metrics} lists popular metrics for evaluating deep CPS models. Section \ref{sec:performance} benchmarks 40 deep CPS models and provides in-depth analyses. Section \ref{sec:challenges} discusses the main challenges and opportunities of deep learning-based CPS methods. Section \ref{sec:conlusion} is the conclusion.

\section{Methodology (Survey)}
\label{sec:method}

This section provides a comprehensive overview of prominent deep CPS methods from novel taxonomy, including network architectures (Sec. \ref{sec:architectures}), level of supervision (Sec. \ref{sec:supervision})and learning paradigm (Sec. \ref{sec:paradigm}), as shown in Fig. \ref{fig:taxonomy}. Table \ref{tab:methods1} and Table \ref{tab:methods2} summarizes the recent proposed deep CPS models and some representative traditional CPS methods. Due to limited space, we only selected a few representative methods for each category.

\subsection{Network Architectures}
\label{sec:architectures}

Based on the adopted backbone architectures, we further classify these deep CPS models into four categories: multilayer perceptron (MLP) based models (Sec. \ref{sec:mlp}), convolutional neural network (CNN) based models (Sec. \ref{sec:cnn}), transformer-based models (Sec. \ref{sec:transformer}),  and hybrid-network based models (Sec. \ref{sec:hybrid}). 

\subsubsection{MLP-based Method}
\label{sec:mlp}
As shown in Fig. \ref{fig:network_structure}(a), MLP is a type of artificial neural network architecture composed of multiple layers of nodes, including an input layer, one or more hidden layers, and an output layer.
Shi \textit{et al.} \cite{shi2022polyp} proposed a novel work to investigate the MLP-based architecture in polyp segmentation, which uses CycleMLP \cite{chen2022cyclemlp} as the encoder to overcome the fixed input scale issue.
Yuan \textit{et al.} \cite{yuan2017automatic} introduce an MLP-based sparse autoencoder to learn high-level superpixel characterization from the hand-crafted features.
Although MLP-based CPS methods exhibit superior performance compared to non-deep learning methods, MLP-based models are limited in leveraging the image's spatial information. Moreover, these methods are time-consuming, as they require processing in a multi-stage manner.

\subsubsection{CNN-based Method}
\label{sec:cnn}

To overcome the limitations of MLP-based methods, recent CPS approaches \cite{shen2022task,tomar2022tganet,wei2022boxpolyp,guo2022non,yang2022source,yue2022boundary,zhao2021automatic,wei2021shallow,tomar2022fanet} adopted the CNN architecture, which enables polyp representation and segmentation in a end-to-end manner. The CNN-based methods have become dominant in the CPS field and can be further divided into FCN-based networks, UNet-based networks, multi-branch networks, and deep supervision networks.

\begin{figure*}[t]
\centering
\resizebox{0.95\textwidth}{!}{
\includegraphics[width=\textwidth]{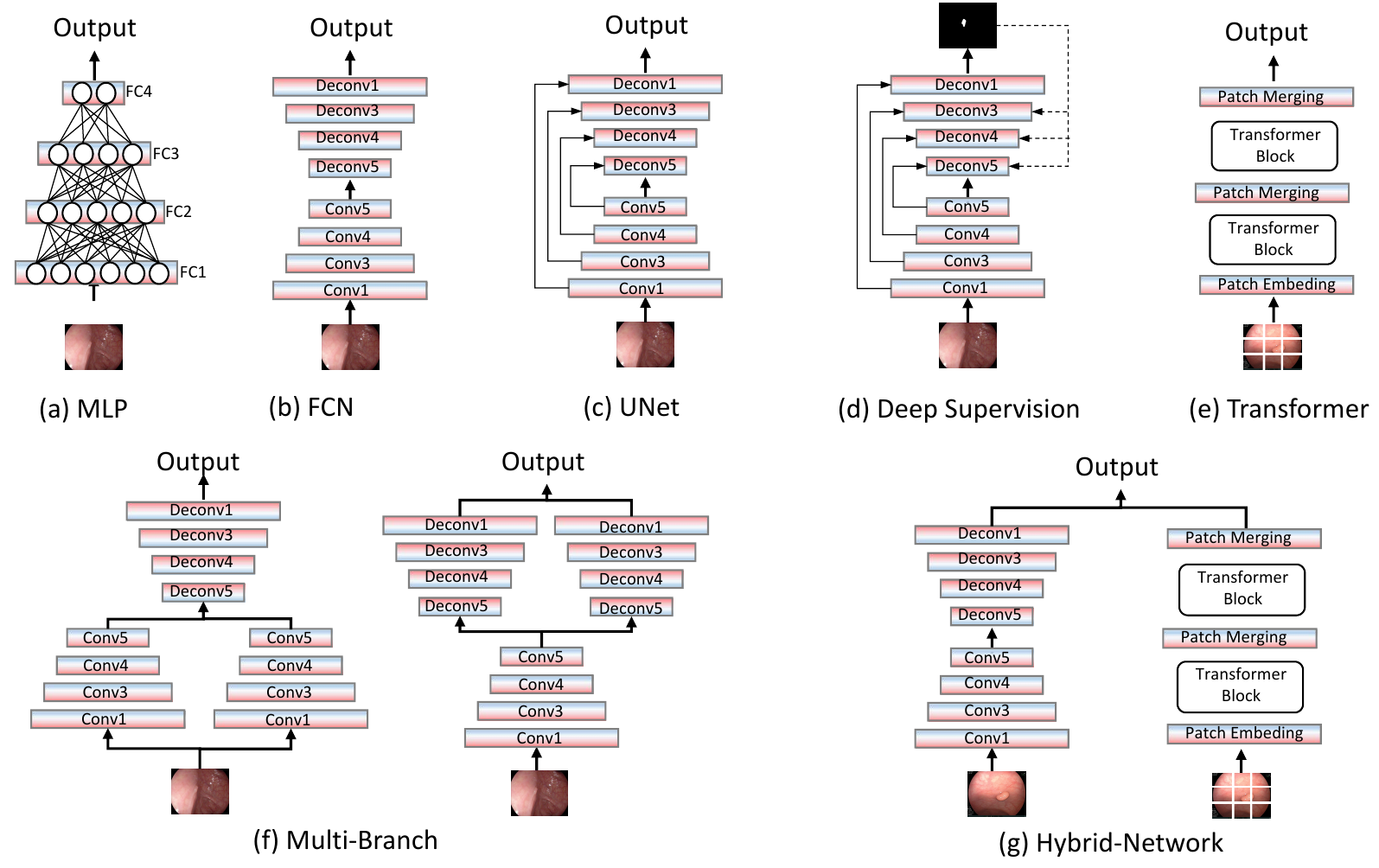}
}
\caption{Classification of deep CPS models according to the network architectures. (a) MLP-based methods, (b) FCN-based models, (c) UNet-based approaches,
 (d) Deep supervision network, (e) Transformer-based network, and (f) Multi-branch networks, and (g) Hybrid-network.}
\label{fig:network_structure}
\end{figure*}

\noindent \textbf{FCN-based Network.} Long \textit{et al.} \cite{long2015fully} introduced Fully Convolutional Networks (FCN) for semantic segmentation, As shown in Fig. \ref{fig:network_structure}(b). Since then, FCN has become the foundational framework for image segmentation, and subsequent CPS methods \cite{nguyen2018colorectal,zhang2017automated,li2017colorectal,brandao2017fully,su2023accurate} largely evolving from it. Yin \textit{et al.} \cite{yin2022duplex} proposed two parallel attention-based modules, which can be incorporated into any encoder-decoder architecture.
Xu \textit{et al.} \cite{xu2022temporal} proposed a temporal correlation network for video polyp segmentation, in which the temporal correlation is unprecedentedly modeled based on the relationship between the original video and the captured frames to be adaptable for video polyp segmentation.
Feng \textit{et al.} \cite{feng2020ssn} proposed a novel stair-shape network for real-time polyp segmentation in colonoscopy images. 
Akbari \textit{et al.} \cite{akbari2018polyp} proposed a polyp segmentation method based on fully the convolutional neural network.
Wichakam \textit{et al.} \cite{wichakam2018real} proposed a compressed fully convolutional network by modifying the FCN-8s network to detect and segment polyp in real-time. 
Wu \textit{et al.} \cite{wu2021precise} proposed a novel ConvNet to accurately segment polyps from colonoscopy videos in a bottom-up/top-down manner.
Dong \textit{et al.} \cite{dong2021asymmetric} present asymmetric attention upsampling, which utilizes the information of low-level feature maps to rescale the high-level feature maps smartly through spatial pooling and attention mechanisms.

\noindent \textbf{UNet-based Network.} Ronneberger \textit{et al.} \cite{ronneberger2015u} proposed the UNet architecture designed for biomedical image segmentation tasks. As shown in Fig. \ref{fig:network_structure}(c), it comprises a contracting path to capture context and a symmetric expanding path, making it suitable for CPS tasks.
For instance, Zhang \textit{et al.} \cite{zhang2022lesion} designed a typical UNet structure, followed by a bottom-up feature extraction and top-down feature fusion strategy to obtain more comprehensive and semantically rich feature representations. 
 Zhao \textit{et al.} \cite{zhao2022semi} a ResUnet based framework to fuse the proximity frame information at different layers and capture contextual information. 
Srivastava \textit{et al.} \cite{srivastava2021msrf}  propose a novel framework for medical image segmentation that consists of an encoder block, a shape stream block, and a decoder block. 
Lin \textit{et al.} \cite{lin2022bsca} proposed a UNet-based network to effectively suppress noises in feature maps and simultaneously improve the ability of feature expression at different levels.
Song \textit{et al.} \cite{song2022attention} designed a new neural network structure based on the currently popular encoding-decoding network architecture.


\noindent \textbf{Multi-branch Network.} 
As shown in Fig. \ref{fig:network_structure}(f), multi-branch-based methods typically consist of multiple encoders/decoders to extract multi-scale features for polyp segmentation explicitly. Li \textit{et al.} \cite{ijcai2022p155} proposed a multi-branch framework, consisting of a segmentation branch and a propagation branch, respectively. Huy \textit{et al.} \cite{huy2022adversarial} proposed an adversarial contrastive Fourier method, learning contrastive loss with supervision from the content encoder and the style encoder. 
Ji \textit{et al.} \cite{ji2022video} design a simple but efficient multistream network (named PNS+) for video polyp segmentation, which consists of a global encoder and a local encoder. The global and local encoders extract long- and short-term spatial-temporal representations from the first anchor frame and multiple successive frames.
Qiu \textit{et al.} \cite{qiu2022bdg} design a two-branch network; one branch is used to generate a boundary distribution map, and another is used to predict the polyp segmentation maps.
Chen \textit{et al.} \cite{chen2022single} proposed a novel two-branched network for polyp segmentation of single-modality endoscopic images, which consists of an image-to-image translation branch and an image segmentation branch.
Tomar \textit{et al.} \cite{tomar2021ddanet} propose a novel deep learning architecture called dual decoder attention network for automatic polyp segmentation.


\noindent \textbf{Deep Supervision Network.} 
As shown in Fig. \ref{fig:network_structure}(d), the deep supervision learning strategy \cite{wu2022msraformer,zhou2018unet,10113675,huang2022transmixer} is proposed to utilize the groundtruth to supervise at each layer of the network, facilitating rapid convergence of the model.
Patel \textit{et al.} \cite{patel2022fuzzynet} proposed a novel fuzzy network to focus more on the difficult pixels, where deep supervision is applied at the end of the output of each fuzzy attention module and partial decoder.
Du \textit{et al.} \cite{du2022icgnet} proposed an integration context-based reverse-contour guidance network with multi-level deep supervision.
Nguyen \textit{et al.} \cite{nguyen2021ccbanet} proposed an encoder-decoder-based architecture and each layer was supervised by the resized ground truth.
Shen \textit{et al.} \cite{shen2021hrenet} proposed a novel hard region enhancement network for polyp segmentation supervised multi-level groundtruth.
\cite{zhang2021transfuse} uses deep supervision to improve the gradient flow by additionally supervising the transformer branch and the first fusion branch.
Kim \textit{et al.} \cite{kim2021uacanet} propose uncertainty augmented context attention network using extra supervision in each layer.
Zhang \textit{et al.} \cite{zhang2020adaptive} propose the adaptive context selection network supervised by the down-sampled groundtruth.
Fan \textit{et al.}  \cite{fan2020pranet} propose a parallel reverse attention network for the polyp segmentation task, adopting deep supervision for the last three side outputs.

\begin{table*}[!htb]
\centering
{
\renewcommand\arraystretch{1.25}
\setlength\tabcolsep{1.5pt}
\resizebox{1\textwidth}{!}{
\begin{tabular}{c|r||cccccccc}
\Xhline{1pt} \rule{0pt}{11pt} {\textbf{Year}} & {\textbf{Methods}} & {\textbf{Publication}}   & {\textbf{Architecture}}    & \textbf{Backbone}     & \textbf{\makecell{Supervision}}      & \textbf{\makecell{Learning Paradigm}} & \textbf{\makecell{Training Dataset}}   & \textbf{\makecell{Number}}                    &  \textbf{Available}                    \\ [3pt] \hline\hline
\multirow{22}{*}{\rotatebox{90}{{\textbf{2023}}}}
           & PETNet \cite{ling2023probabilistic}  & MICCAI   & Transformer   & PVTv2-B2 \cite{wang2022pvt} & Fully-Sup.    & Gaussian-Probabilistic, Ensemble& ClinicDB \cite{bernal2015wm}+Kvasir-SEG \cite{jha2020kvasir}  & 548+900        &  No \\

           & RPFA \cite{su2023revisiting} & MICCAI  & Transformer+UNet   & PVT \cite{wang2021pyramid}  & Fully-Sup.    & Feature Propagation/Fusion & ClinicDB \cite{bernal2015wm}+Kvasir-SEG \cite{jha2020kvasir}   & 548+900       &  No \\

           & S2ME \cite{wang2023s2me} & MICCAI   & CNN+UNet   &ResNet50 \cite{he2016deep}  &  Scribble-Sup. & Ensemble Learning, Mutual Teaching&  SUN-SEG\cite{ji2022video}   & 6677        & \href{https://github.com/lofrienger/S2ME}{Open}  \\

           & WeakPolpy \cite{wei2023weakpolyp} & MICCAI   & \makecell{Transformer/FCN}   & \makecell{Res2Net50\cite{gao2019res2net}/PVTv2-B2 \cite{wang2022pvt} }  & Box-Sup.   & Transformation, Scale Consistency& \makecell{SUN-SEG\cite{ji2022video}/POLYP-SEG\cite{wei2023weakpolyp}}   & \makecell{19,544/15,916}       & \href{https://github.com/weijun88/WeakPolyp}{Open}  \\

           & FSFM \cite{su2023accurate}  & ISBI   & CNN+FPN   & VAN \cite{guo2023visual}   & Fully-Sup.    &Feature Fusion, Attention & ClinicDB \cite{bernal2015wm}+Kvasir-SEG \cite{jha2020kvasir}   & 548+900        & No  \\

           & EPSG \cite{haithami2023enhancing}  & ISBI   & CNN+UNet   & ResNet50 \cite{he2016deep}  & Fully-Sup.    & Image Transformation, Variation & ClinicDB \cite{bernal2015wm}+Kvasir-SEG \cite{jha2020kvasir}  & 490+800      & No  \\ 

           & RealSeg \cite{su2023go} & ISBI & FPN & VAN \cite{guo2023visual} &  Fully-Sup.& Real-time, Feature Fusion & ClinicDB \cite{bernal2015wm}+Kvasir-SEG \cite{jha2020kvasir}   & 548+900        & No  \\ 

           & PSTNet \cite{xiong2023unpaired} & ISBI & CNN+UNet & Res2Net50\cite{gao2019res2net} & Fully-Sup. &Domain Adaptation, Translation& PICCOLO \cite{app10238501} & 3,546 & No \\

           & EMTSNet \cite{wang2023efficient} & JBHI & CNN+UNet & Res2Net50\cite{gao2019res2net} & Fully-Sup. & Multi-Task, Activation Map& ClinicDB \cite{bernal2015wm}+Kvasir-SEG \cite{jha2020kvasir}   & 548+900        & No\\ 

           &GAN-PSNet \cite{10302215} & JBHI & GAN+UNet & Res2Net50\cite{gao2019res2net} & Fully-Sup. & Attention, Generative Adversarial & Kvasir-SEG \cite{jha2020kvasir}   & 800        & No\\ 

           & FEGNet \cite{10113675} & JBHI & CNN+UNet & Res2Net50\cite{gao2019res2net} & Fully-Sup. &Feedback Mechanism, Boundary & ClinicDB \cite{bernal2015wm}+Kvasir-SEG \cite{jha2020kvasir}   & 548+800        & No  \\

           & PMSACL \cite{tian2023self} & MIA & CNN+UNet & ResNet18\cite{he2016deep} & Self/Un-Sup. & \makecell{Data augmentation, Anomaly} &HyperKvasir \cite{Borgli2020} & 2100 & No \\

           & XBFormer \cite{wang2023xbound} & TMI& Transformer & PVT \cite{wang2021pyramid}  & Fully-Sup. & Cross-scale, Boundary Fusion & ClinicDB \cite{bernal2015wm}+Kvasir-SEG \cite{jha2020kvasir}   & 548+800        & \href{https://github.com/jcwang123/xboundformer}{Open}  \\

           & ColnNet \cite{jain2023coinnet} & TMI & CNN+UNet & DenseNet121 \cite{huang2017densely} & Fully-Sup. &Relationship of Feature, Boundary& ClinicDB \cite{bernal2015wm}+Kvasir-SEG \cite{jha2020kvasir}   & 548+800        & No  \\

           & RPANet \cite{wang2023unsupervised} & IPMI & CNN+UNet & ResNet101 \cite{he2016deep} & \makecell{Fully-Sup. } & Domain Adaptation, Self-Supervision &Private Dataset \cite{wang2023unsupervised}   & 5,175     & No  \\

           & TransNetR \cite{jha2023transnetr} & MIDL & Transformer+CNN & ResNet50 \cite{he2016deep} & Fully-Sup. & Out-of-distribution, Generalization & Kvasir-SEG \cite{jha2020kvasir} & 900 & \href{https://github.com/DebeshJha/TransNetR}{Open} \\

           &CFANet \cite{zhou2023cross} & PR & CNN+UNet & Res2Net50\cite{gao2019res2net} &  Fully-Sup. & Boundary-aware, Feature Fusion & ClinicDB \cite{bernal2015wm}+Kvasir-SEG \cite{jha2020kvasir}   & 548+900  & \href{https://github.com/taozh2017/CFANet}{Open} \\

           & GSAL \cite{wang2023gsal} & PR & CNN+GAN & VGG16 \cite{simonyan2015very}&  Fully-Sup. &Adversarial Learning & ClinicDB \cite{bernal2015wm}+Kvasir-SEG \cite{jha2020kvasir}   & 548+900        & \href{https://github.com/DLWK/GSAL}{Open} \\

           & Polyp-Mixer \cite{shi2022polyp} & TCSVT & MLP & CycleMLP-B1 \cite{chen2022cyclemlp} &  Fully-Sup. & Context-Aware, MLP& ClinicDB \cite{bernal2015wm}+Kvasir-SEG \cite{jha2020kvasir}   & 548+900        & No \\

           & PTMFNet \cite{wang2023pyramid} & ICIP & Transformer & PVTv2 \cite{wang2022pvt} & Fully-Sup. &Pyramid Transformer, Multibranch& -  & -        & No \\

           & PLCUT-Seg \cite{moreu2023self} & IJCNN & CNN & HarDNet68 \cite{chao2019hardnet} & Self/Semi-Sup. &Synthetic Data, Self/Semi-Supervised& ClinicDB \cite{bernal2015wm}+Kvasir-SEG \cite{jha2020kvasir}   & 548+900        & No \\

           & PolypSeg+ \cite{wu2022polypseg} & TCYB & CNN+UNet & ResNet50 \cite{he2016deep} & Fully-Sup.& Context-aware, Feature Fusion& ClinicDB \cite{bernal2015wm}+Kvasir-SEG \cite{jha2020kvasir}   & 548+900        & No \\

           & CASCADE \cite{rahman2023medical} & WACV & Transformer+UNet & PVTv2 \cite{wang2022pvt} &  Fully-Sup.& Cascaded Attention, Feature Fusion& \makecell{\href{https://www.synapse.org/\#!Synapse:syn3193805/wiki/217789}{Synapse}+ACDC\cite{bernard2018deep} +\\ClinicDB \cite{bernal2015wm}+Kvasir-SEG \cite{jha2020kvasir}}    & \makecell{2212+1930\\+548+900}        & \href{https://github.com/SLDGroup/CASCADE}{Open} \\

           & PolypPVT \cite{dong2023PolypPVT} & CAAI AIR & Transformer+UNet & PVT\cite{wang2021pyramid} & Fully-Sup. & \makecell{Cascaded Fusion} & ClinicDB \cite{bernal2015wm}+Kvasir-SEG \cite{jha2020kvasir} & 550+900  & \href{https://github.com/DengPingFan/Polyp-PVT}{Open} \\

           & PEFNet \cite{nguyen2023pefnet} & MMM & CNN+UNet & EfficientNetV2-L \cite{tan2021efficientnetv2} & Fully-Sup.& Positional Embedding & Kvasir-SEG \cite{jha2020kvasir}   & 600        & \href{https://github.com/huyquoctrinh/PEFNet}{Open} \\ \hline \hline

\multirow{35}{*}{\rotatebox{90}{{\textbf{2022}}}}

           & TRFR-Net \cite{shen2022task} & MICCAI & CNN+GAN & ResNet34 \cite{he2016deep} & Fully-Sup.& Domain Adaptation, Adversarial Learning & \makecell{Kvasir-SEG \cite{jha2020kvasir}+ETIS-Larib \cite{silva2014toward} \\+ CVC-ColonDB \cite{tajbakhsh2015automated} }  & 700+137+210       & No \\

           & TGANet \cite{tomar2022tganet} & MICCAI & CNN+UNet & ResNet50 \cite{he2016deep} & Fully-Sup. & \makecell{Attention, Multi-scale} & Kvasir-SEG \cite{jha2020kvasir}   & 880  & \href{https://github.com/nikhilroxtomar/tganet}{Open} \\

           &LDNet \cite{zhang2022lesion} & MICCAI & CNN+UNet & Res2Net50\cite{gao2019res2net} & Fully-Sup. & \makecell{Dynamci Kernel, Attention} & ClinicDB \cite{bernal2015wm}+Kvasir-SEG \cite{jha2020kvasir} & 548+800 & \href{https://github.com/ReaFly/LDNet}{Open} \\

           & SSFormer \cite{wang2022stepwise} & MICCAI & Transformer+UNet & PVTv2 \cite{wang2022pvt} &  Fully-Sup.&  \makecell{Local,Global Fusion} & ClinicDB \cite{bernal2015wm}+Kvasir-SEG \cite{jha2020kvasir} & 548+900 & \href{https://github.com/Qiming-Huang/ssformer}{Open} \\

           & BoxPolyp \cite{wei2022boxpolyp} & MICCAI & Transformer/CNN  &\makecell{PVTv2 \cite{wang2022pvt}/Res2Net50\cite{gao2019res2net} } & Box-Sup.&  \makecell{Fusion,  Consistency Loss} & ClinicDB \cite{bernal2015wm}+Kvasir-SEG \cite{jha2020kvasir} & 548+900 & \href{https://github.com/weijun88/BoxPolyp}{Open} \\

           & PPFormer \cite{cai2022using} & MICCAI & Transformer+CNN & CvT\cite{wu2021cvt}+VGG16\cite{simonyan2015very} & Fully-Sup.&  \makecell{Self-attention, Local2Global Fusion} &  ClinicDB \cite{bernal2015wm}+Kvasir-SEG \cite{jha2020kvasir} & 548+900 & No \\

           & SSTAN \cite{zhao2022semi} & MICCAI & CNN+UNet & ResNet34\cite{he2016deep} & Semi-Sup. & \makecell{Context Attention} & LDPolypVideo \cite{ma2021ldpolypvideo} & 693+14704 & No \\

            & NIP \cite{guo2022non} & MedIA & CNN+UNet & ResNet101\cite{he2016deep} &  Fully-Sup. & \makecell{Data Augmentation, Resampling} & EndoScene \cite{vazquez2017benchmark} & 547 & No \\

           & FSM \cite{yang2022source} & MedIA & CNN+UNet & ResNet101\cite{he2016deep} &  Fully-Sup. & \makecell{Domain Adaptation, Resampling} & EndoScene \cite{vazquez2017benchmark} & 547 & \href{https://github.com/CityU-AIM-Group/SFDA-FSM}{Open} \\

           & MSRF-Net \cite{srivastava2021msrf} & JBHI & CNN+UNet & SENet \cite{hu2018squeeze} & Fully-Sup.& \makecell{Multi-scale Fusion} &  ClinicDB \cite{bernal2015wm}+Kvasir-SEG \cite{jha2020kvasir} & 490+800 & \href{https://github.com/NoviceMAn-prog/MSRF-Net}{Open} \\

           &BCNet \cite{yue2022boundary} & JBHI & CNN+UNet & Res2Net50\cite{gao2019res2net}  & Fully-Sup.& \makecell{Cross-Feature Fusion, Boundary} &  Kvasir-SEG \cite{jha2020kvasir} & 800 & \href{https://github.com/NoviceMAn-prog/MSRF-Net}{Open} \\

           & PNS+ \cite{ji2022video} & MIR & CNN+UNet & Res2Net50\cite{gao2019res2net}  & Fully-Sup. & \makecell{Feature Aggregation, Attention} & SUN-SEG \cite{ji2022video} & 19,544 & \href{https://github.com/GewelsJI/VPS}{Open} \\

           & FCBFormer \cite{sanderson2022fcn} & MIUA & Transformer+CNN & \makecell{PVTv2-B3 \cite{wang2022pvt}+ResNet34\cite{he2016deep}} & Fully-Sup. &Two-stream Fusion & ClinicDB \cite{bernal2015wm}+Kvasir-SEG \cite{jha2020kvasir} & 490+800 & \href{https://github.com/ESandML/FCBFormer}{Open} \\

           & BDGNet \cite{qiu2022bdg} & SPIE MI & CNN+UNet & EfficientNet-B5 \cite{tan2019efficientnet} & Fully-Sup. & \makecell{Multi-Scale Fusion, Boundary} &  ClinicDB \cite{bernal2015wm}+Kvasir-SEG \cite{jha2020kvasir} & 550+900  & \href{https://github.com/zihuanqiu/BDG-Net}{Open} \\

           & TCNet \cite{xu2022temporal} & BIBM & FCN & Res2Net50\cite{gao2019res2net} &  Fully-Sup. & \makecell{Temporal Correlation Modeling} & ClinicDB \cite{bernal2015wm}+Kvasir-SEG \cite{jha2020kvasir} & 550+800  & No \\

           & TransMixer \cite{huang2022transmixer} & BIBM & Transformer+CNN& PVTv2 \cite{wang2022pvt}+SENet \cite{hu2018squeeze}& Fully-Sup. & \makecell{Interaction Fusion, Attention} & ClinicDB \cite{bernal2015wm}+Kvasir-SEG \cite{jha2020kvasir} & 550+900  & No \\

           & ICBNet \cite{xiao2022icbnet} & BIBM & Transformer+UNet & PVT \cite{wang2021pyramid} & Fully-Sup. & \makecell{Iterative Fusion, Attention} & ClinicDB \cite{bernal2015wm}+Kvasir-SEG \cite{jha2020kvasir} & 550+900  & No \\

           & CLDNet \cite{chen2022cld} & BIBM &  Transformer+UNet & PVTv2 \cite{wang2022pvt} & Fully-Sup. & \makecell{Multi-Scale Fusion, Boundary} & ClinicDB \cite{bernal2015wm}+Kvasir-SEG \cite{jha2020kvasir} & 490+800  & No \\

           & TASNet \cite{chen2022single} & BIBM & CNN+UNet & Res2Net50\cite{gao2019res2net}  & Fully-Sup. & \makecell{Color Reversal, Two-Branched Network} & ClinicDB \cite{bernal2015wm}+Kvasir-SEG \cite{jha2020kvasir} & 550+900  & No \\

           & FuzzyNet \cite{patel2022fuzzynet} & NeurIPS& CNN/Transformer & Res2Net50\cite{gao2019res2net}/PVT \cite{wang2021pyramid} &  Fully-Sup. & \makecell{Fuzzy Attention} & ClinicDB \cite{bernal2015wm}+Kvasir-SEG \cite{jha2020kvasir} & 550+900  & \href{https://github.com/krushi1992/FuzzyNet}{Open} \\

           & SwinPA-Net \cite{du2022swinpa} & TNNLS & Transformer+UNet & Swin-B \cite{liu2021swin} &  Fully-Sup. & \makecell{Pyramid Attention, Multiscale Fusion} & ClinicDB \cite{bernal2015wm}+Kvasir-SEG \cite{jha2020kvasir} & 550+900  & No \\

           & BSCA-Net \cite{lin2022bsca} & PR & CNN+UNet & Res2Net50\cite{gao2019res2net} & Fully-Sup. & \makecell{Multipath, Attention} & ClinicDB \cite{bernal2015wm}+Kvasir-SEG \cite{jha2020kvasir} & 550+900  & No \\

           & HSNet \cite{zhang2022hsnet} & CBM & CNN+Transformer &  PVTv2 \cite{wang2022pvt}+ Res2Net50\cite{gao2019res2net} & Fully-Sup. & \makecell{Cross Attention, Semantic Complementary} & ClinicDB \cite{bernal2015wm}+Kvasir-SEG \cite{jha2020kvasir} & 548+900  & \href{https://github.com/baiboat/HSNet}{Open} \\

           & MSRAformer \cite{wu2022msraformer} & CBM & Transformer+UNet & Swin-B \cite{liu2021swin}  & Fully-Sup. & \makecell{Channel/Reverse Attention} & ClinicDB \cite{bernal2015wm}+Kvasir-SEG \cite{jha2020kvasir} & 490+800  & \href{https://github.com/ChengLong1222/MSRAformer-main}{Open} \\

           & AMNet \cite{song2022attention} & CBM & CNN+UNet & Res2Net50\cite{gao2019res2net} & Fully-Sup. & \makecell{Multiscale Fusion, Attention} & ClinicDB \cite{bernal2015wm}+Kvasir-SEG \cite{jha2020kvasir} & 500+800  & No \\

           & DBMF \cite{liu2022dbmf} & CBM & Transformer+CNN & Swin-B \cite{liu2021swin}+EfficientNet\cite{tan2019efficientnet} & Fully-Sup. & \makecell{Multiscale Fusion, Attention} & ClinicDB \cite{bernal2015wm}+Kvasir-SEG \cite{jha2020kvasir} & 550+900  & No \\

           & ICGNet \cite{du2022icgnet} & IJCAI & CNN+UNet & ResNet34\cite{he2016deep} & Fully-Sup. & \makecell{Local-Global Fusion, Boundary Supervision} & EndoScene \cite{vazquez2017benchmark}+Kvasir-SEG \cite{jha2020kvasir} & 547+600  & No \\

           & TCCNet \cite{ijcai2022p155} & IJCAI & CNN+UNet & Res2Net50\cite{gao2019res2net} & Semi-Sup. & \makecell{ Context-Free Loss,  Reverse Attention} &ClinicDB \cite{bernal2015wm}+CVC-ColonDB \cite{tajbakhsh2015automated} & 367+180  & \href{https://github.com/wener-yung/TCCNet}{Open} \\

           & CoFo \cite{huy2022adversarial} & ISBI & CNN+UNet & ResNet18\cite{he2016deep} & Fully-Sup. & \makecell{Domain Adapation, Adversarial Learning} & EndoScene \cite{vazquez2017benchmark}+Kvasir-SEG \cite{jha2020kvasir} & 547+600  & \href{https://github.com/tadeephuy/CoFo}{Open} \\

           & DCRNet \cite{yin2022duplex} & ISBI & CNN+UNet & ResNet34\cite{he2016deep} & Fully-Sup. & \makecell{Contextual-Relation Learning} & \makecell{EndoScene \cite{vazquez2017benchmark}+Kvasir-SEG \cite{jha2020kvasir} \\+ Piccolo\cite{sanchez2020piccolo}} & \makecell{547+600\\+2203}  & \href{https://github.com/tadeephuy/CoFo}{Open} \\ \hline \hline

\multirow{24}{*}{\rotatebox{90}{{\textbf{2021}}}}

             & MSNet \cite{zhao2021automatic} & MICCAI & CNN+UNet & Res2Net50\cite{gao2019res2net} & Fully-Sup. & \makecell{Multiscale Fusion, Loss Network} & ClinicDB \cite{bernal2015wm}+Kvasir-SEG \cite{jha2020kvasir} & 550+900  & \href{https://github.com/Xiaoqi-Zhao-DLUT/MSNet-M2SNet}{Open} \\

             & CCBANet \cite{nguyen2021ccbanet} & MICCAI & CNN+UNet & ResNet34\cite{he2016deep} & Fully-Sup. & \makecell{Cascading Context, Attention} & ClinicDB \cite{bernal2015wm}+Kvasir-SEG \cite{jha2020kvasir} & 550+900  & \href{https://github.com/ntcongvn/CCBANet}{Open} \\

             &HRNet \cite{shen2021hrenet} & MICCAI &  CNN+UNet & ResNet34\cite{he2016deep} & Fully-Sup. & \makecell{Feature Aggregation, Attention} & ClinicDB \cite{bernal2015wm}+Kvasir-SEG \cite{jha2020kvasir} & 550+900  & No \\

             &LODNet \cite{cheng2021learnable} & MICCAI & CNN & ResNet50\cite{he2016deep} & Fully-Sup. & \makecell{Oriented, Sensitive Loss} & ClinicDB \cite{bernal2015wm}+Kvasir-SEG \cite{jha2020kvasir} & 550+900  & \href{https://github.com/midsdsy/LOD-Net}{Open} \\


             &SANet \cite{wei2021shallow} & MICCAI & CNN+UNet & Res2Net50\cite{gao2019res2net} & Fully-Sup. & \makecell{Color Exchange, Attention} & ClinicDB \cite{bernal2015wm}+Kvasir-SEG \cite{jha2020kvasir} & 550+900  & \href{https://github.com/weijun88/SANet}{Open} \\

             & Transfuse \cite{zhang2021transfuse} & MICCAI & Transformer+CNN & Res2Net34\cite{gao2019res2net}+DeiT-S\cite{touvron2021training} & Fully-Sup. & \makecell{ Feature Fusion} & ClinicDB \cite{bernal2015wm}+Kvasir-SEG \cite{jha2020kvasir} & 550+900  & \href{https://github.com/Rayicer/TransFuse}{Open} \\

              & CCD \cite{tian2021constrained} & MICCAI & CNN+UNet& ResNet18\cite{he2016deep} & Self/Un-Sup. & \makecell{Contrastive Learning, Anomaly Localisation} & HyperKvasir \cite{Borgli2020} & 2100 & \href{https://github.com/tianyu0207/CCD}{Open} \\

             & Polyformer \cite{li2021few} & MICCAI & Transformer+UNet& Segtran\cite{li576medical} & Fully-Sup. & \makecell{Few-Shot Learning, Domain Adaptation} & ClinicDB \cite{bernal2015wm}+Kvasir-SEG \cite{jha2020kvasir} & 550+900 & \href{https://github.com/askerlee/segtran}{Open} \\

             &PNet \cite{ji2021progressively} & MICCAI & CNN &  Res2Net50\cite{gao2019res2net} & Fully-Sup. & \makecell{Feature Aggregation, Graph Convolution} &ClinicDB \cite{bernal2015wm}+Kvasir-SEG \cite{jha2020kvasir} & 550+900 & \href{https://github.com/DengPingFan/PraNet}{Open} \\

              & HieraSeg \cite{guo2021dynamic} & MIA & CNN+FCN & Deeplabv3+\cite{chen2017rethinking}  & Fully-Sup. & \makecell{Hierarchical, Dynamic Weighting} & EndoScene\cite{vazquez2017benchmark} & 547 & \href{https://github.com/CityU-AIM-Group/DW-HieraSeg}{Open} \\

              & ThresholdNet \cite{guo2020learn} & TMI & CNN+FCN & Deeplabv3+\cite{chen2017rethinking} & Fully-Sup. & \makecell{Mixup, Threshold Loss} &EndoScene\cite{vazquez2017benchmark} & 547 & \href{https://github.com/Guo-Xiaoqing/ThresholdNetv}{Open} \\

              &FANet \cite{tomar2022fanet} & TNNLS &  CNN+UNet  & N/A & Fully-Sup. & \makecell{Iterative Refining, Feedback Attention} & ClinicDB \cite{bernal2015wm}+Kvasir-SEG \cite{jha2020kvasir} & 550+900  & \href{https://github.com/nikhilroxtomar/FANet}{Open} \\

             & MPA-DA \cite{yang2021mutual} & JBHI & CNN+UNet & ResNet101\cite{he2016deep} & Fully-Sup. & \makecell{Domain Adaptation, Self-training} & ClinicDB \cite{bernal2015wm}+Kvasir-SEG \cite{jha2020kvasir} & 548+1000  & \href{https://github.com/CityU-AIM-Group/MPA-DA}{Open} \\

             & ResUNet++\cite{jha2021comprehensive} & JBHI & CNN+UNet & ResUNet\cite{zhang2018road} & Fully-Sup. & \makecell{Test Time Augmentation} & ClinicDB \cite{bernal2015wm}+Kvasir-SEG \cite{jha2020kvasir} & 550+900  & \href{https://github.com/DebeshJha/ResUNetPlusPlus-with-CRF-and-TTA}{Open} \\

             & SCRNet \cite{wu2021precise} & AAAI & CNN+UNet & FPN\cite{lin2017feature} & Fully-Sup. & \makecell{Semantic Calibration/Refinement} &Kvasir-SEG \cite{jha2020kvasir} & 700 & No \\

             & CAFD \cite{wu2021collaborative} & ICCV & CNN+UNet & ResNet50\cite{he2016deep} & Semi-Sup. & \makecell{Adversarial Learning, Collaborative Learning} &ClinicDB \cite{bernal2015wm}+Kvasir-SEG \cite{jha2020kvasir} & 306+500 & No \\

             & UACANet \cite{kim2021uacanet} & ACM MM & CNN+UNet & Res2Net\cite{gao2019res2net} & Fully-Sup. & \makecell{Uncertainty, Self-attention} & ClinicDB \cite{bernal2015wm}+Kvasir-SEG \cite{jha2020kvasir} & 550+900  & \href{https://github.com/plemeri/UACANet}{Open} \\

              & Segtran\cite{li576medical} & IJCAI & Transformer+CNN & ResNet101\cite{he2016deep}  & Fully-Sup. & \makecell{Squeezed Attention,  Positional Encoding} &ClinicDB \cite{bernal2015wm}+Kvasir-SEG \cite{jha2020kvasir} & 489+800 & \href{https://github.com/askerlee/segtran}{Open} \\

             & DenseUNet \cite{safarov2021denseunet} & Sensors & CNN+UNet  & DenseNet\cite{huang2017densely} & Fully-Sup. & \makecell{Dense Connections, Attention} & ClinicDB \cite{bernal2015wm}+Kvasir-SEG \cite{jha2020kvasir} & 490+800  & No \\

             &EUNet \cite{patel2021enhanced} & CVR & CNN+UNet & ResNet34\cite{he2016deep} & Fully-Sup. & \makecell{Semantic Enhancement, Global Context} & ClinicDB \cite{bernal2015wm}+Kvasir-SEG \cite{jha2020kvasir} & 550+900  & \href{https://github.com/rucv/Enhanced-U-Net}{Open} \\

             &DDANet \cite{tomar2021ddanet} & ICPR & CNN+UNet & ResUNet\cite{zhang2018road} & Fully-Sup. & \makecell{Dual Decoder} &Kvasir-SEG \cite{jha2020kvasir} & 880  & \href{https://github.com/nikhilroxtomar/DDANet}{Open} \\

             & FUNet \cite{yeung2021focus} & CBM &  CNN+UNet & ResNet34\cite{he2016deep} & Fully-Sup. & \makecell{Dual Attention, Gate} &ClinicDB \cite{bernal2015wm}+Kvasir-SEG \cite{jha2020kvasir} & 550+900 & No \\

             &T-UNet \cite{tran2021tmd} & Healthcare & CNN+UNet & ResNet50\cite{he2016deep} & Fully-Sup. & \makecell{Dilated Convolution, Feature Fusion} &ClinicDB \cite{bernal2015wm}& 489 & No \\

             & DNets\cite{divergentNets} & ISBI & CNN+FCN & FPN\cite{lin2017feature}+Deeplabv3+\cite{chen2017rethinking}  & Fully-Sup. & \makecell{Ensemble Learning} &ClinicDB \cite{bernal2015wm}+Kvasir-SEG \cite{jha2020kvasir} & 489+800 & \href{https://github.com/vlbthambawita/divergent-nets}{Open} \\

             & AAU \cite{dong2021asymmetric} & ISBI & CNN+FCN & ResNet34\cite{he2016deep}  & Fully-Sup. & \makecell{ Asymmetric Attention, Upsampling} &Kvasir-SEG \cite{jha2020kvasir} & 800 & No \\

              & BI-GCN \cite{meng2021bi} & BMVC & GCN &  Res2Net50\cite{gao2019res2net} & Fully-Sup. & \makecell{Feature Aggregation, Graph Convolution} &ClinicDB \cite{bernal2015wm}+Kvasir-SEG \cite{jha2020kvasir} & 550+900 & \href{https://github.com/smallmax00/BI-GConv}{Open} \\

           \Xhline{0.6pt}

\end{tabular}
}}
\caption{Summary of popular deep CPS methods from 2021 to 2023, including network architecture, backbone, level of supervision, learning paradigm, training dataset and the source code. Section \ref{sec:method} gives more detailed descriptions.}
\label{tab:methods1}
\end{table*} 

\begin{table*}[!htb]
\centering
{
\renewcommand\arraystretch{1.2}
\setlength\tabcolsep{1.5pt}
\resizebox{1\textwidth}{!}{
\begin{tabular}{c|r||cccccccc}
\Xhline{1pt} \rule{0pt}{11pt} {\textbf{Year}} & {\textbf{Methods}} & {\textbf{Publication}}   & {\textbf{Architecture}}    & \textbf{Backbone}     & \textbf{\makecell{Supervision}}      & \textbf{\makecell{Learning Paradigm}} & \textbf{\makecell{Training Dataset}}   & \textbf{\makecell{Number}}                    &  \textbf{Available}                    \\ [3pt] \hline\hline
\multirow{10}{*}{\rotatebox{90}{{\textbf{2020}}}}

           &ACSNet \cite{zhang2020adaptive} & MICCAI & CNN+UNet & ResNet34 \cite{he2016deep} &  Fully-Sup.  & Adaptive Selection, Attention & ClinicDB \cite{bernal2015wm}+Kvasir-SEG \cite{jha2020kvasir}  & 547+600        &  \href{https://github.com/ReaFly/ACSNet}{Open} \\

           & MI$^2$GAN \cite{xie2020mi} & MICCAI & GAN & CycleGAN  \cite{zhu2017unpaired} &  Fully-Sup.  & Domain Adaptation, Adversarial Learning & ClinicDB \cite{bernal2015wm} & 490       &  No \\

           & PraNet \cite{fan2020pranet} & MICCAI & CNN+UNet & Res2Net50\cite{gao2019res2net} &  Fully-Sup.  & Feature Aggregating, Attention & ClinicDB \cite{bernal2015wm}+Kvasir-SEG \cite{jha2020kvasir}  & 547+800    &  \href{https://github.com/DengPingFan/PraNet}{Open} \\

           & Polypseg \cite{zhong2020polypseg} & MICCAI & CNN+UNet & UNet \cite{ronneberger2015u} & Fully-Sup.  &Context-aware, Semantic& Kvasir-SEG \cite{jha2020kvasir}  & 600    &  No \\

           & ABCNet \cite{fang2020abc} & Sensors & CNN+FCN & ResNeXt50 \cite{xie2017aggregated} & Fully-Sup.  &Boundary Loss-aware, Two-stream& Kvasir-SEG \cite{jha2020kvasir}+EndoScene\cite{vazquez2017benchmark}  & 800+547  &  No \\

           &UINet  \cite{wickstrom2020uncertainty} & MIA & CNN & Res2Net50\cite{gao2019res2net} &  Fully-Sup.  & Uncertainty, {Interpretability} & EndoScene\cite{vazquez2017benchmark} & 547  & No \\

           & MCNet \cite{wang2020multi} & JBHI & CNN+UNet & VGG16\cite{simonyan2015very} & Fully-Sup. & Global Semantic, Local Detail & ClinicDB \cite{bernal2015wm}+\href{https://endovissub-abnormal.grand-challenge.org/}{EndoVC}  & 412+465  & No \\

           & SSN \cite{feng2020ssn} & ISBI & CNN+FCN & ResNet \cite{he2016deep} & Fully-Sup. &Multi-scale Fusion, Attention & ClinicDB \cite{bernal2015wm}+Kvasir-SEG \cite{jha2020kvasir}  & 547+800  & No \\

           & RCIS \cite{huang2020real} & ICARM & CNN+FCN & ResNet \cite{he2016deep} & Fully-Sup. &Knowledge Distillation & ClinicDB \cite{bernal2015wm}+Kvasir-SEG \cite{jha2020kvasir}  & 547+800  & No \\

           &Double-UNet \cite{jha2020doubleu} & CBMS & CNN+UNet & VGG19\cite{simonyan2015very} & Fully-Sup. & Multi-branch, Feature Fusion & ClinicDB \cite{bernal2015wm}+ETIS-Larib \cite{silva2014toward}   & 490+157  & No \\ \hline \hline

\multirow{10}{*}{\rotatebox{90}{{\textbf{2019}}}}

          &SFANet \cite{fang2019selective} & MICCAI & CNN+UNet & ResNet34 \cite{he2016deep} & Fully-Sup. & Selective Aggregation, Boundary Loss & EndoScene\cite{vazquez2017benchmark} & 730  & No \\ 

          & TDE \cite{de2019training} & CBMS & CNN+UNet & ResNet34 \cite{he2016deep} & Fully-Sup. & Data Augmentation & ClinicDB \cite{bernal2015wm} & 612  & No \\

          & PsiNet \cite{murugesan2019psi} & EMBC & CNN+UNet & ResNet50 \cite{he2016deep} & Fully-Sup. & Boundary, Multi-task Learning & EndoScene\cite{vazquez2017benchmark} & 638  & \href{https://github.com/Bala93/Multi-task-deep-network}{Open} \\

          &  PSGAN \cite{poorneshwaran2019polyp} & EMBC & GAN & ResNet50 \cite{he2016deep} & Fully-Sup. & Generative Adversarial Network & ClinicDB \cite{bernal2015wm} & 488  & No \\

          & CPSUNet\cite{sun2019colorectal} & ICMLA & CNN+UNet & ResNet50 \cite{he2016deep} & Fully-Sup. & Dilated Convolution, Morphological & ClinicDB \cite{bernal2015wm}+ GIANA\cite{bernal2017comparative} & 300+56  & No \\ 

          & ResUNet++ \cite{jha2019resunet} & ISM & CNN+UNet & ResUNet \cite{zhang2018road} & Fully-Sup. & Squeeze-Excitation, Attention & ClinicDB \cite{bernal2015wm}+Kvasir-SEG \cite{jha2020kvasir}  & 547+800 & \href{https://github.com/DebeshJha/ResUNetPlusPlus-with-CRF-and-TTA}{Open} \\ 

          & PDS \cite{qadir2019polyp} & ISMICT & CNN & Maks R-CNN \cite{he2017mask} & Fully-Sup. & Feature Extractor, Ensemble & ClinicDB \cite{bernal2015wm}+ ETIS-Larib\cite{silva2014toward}  & 547+157 & No \\ 

          & GIANA \cite{guo2019giana} & IJCCV & CNN+FCN &  ResNet50 \cite{he2016deep} &  Fully-Sup. & Dilated Convolution, Squeeze-Excitation &  ClinicDB \cite{bernal2015wm}+ GIANA\cite{bernal2017comparative} & 300+56  & No \\ 

          & CCS \cite{bagheri2019deep} & EMBC & CNN & LinkNet \cite{chaurasia2017linknet} &  Fully-Sup. & Color Space Transformations & CVC-ColonDB \cite{tajbakhsh2015automated}  & 284  & No \\ 

          & APS \cite{guo2019automated} & Med Phys& CNN+UNet & VGG16 \cite{simonyan2015very} &  Fully-Sup. & Data Augmentation, Ensemble Learning &  ClinicDB \cite{bernal2015wm}+ ETIS-Larib\cite{silva2014toward}  & 547+157  & No \\

          \hline \hline

\multirow{14}{*}{\rotatebox{90}{{\textbf{2014-2018}}}}

          & PSFCN \cite{akbari2018polyp} & EMBC & CNN+FCN & FCN-8S \cite{long2015fully} &  Fully-Sup. & Patch Selection, Data Augmentation  &  CVC-ColonDB \cite{tajbakhsh2015automated} & 200  & No \\ 

          & ACPS \cite{sanchez2018automatized} & CBM & N/A  &  N/A  &  Un-Sup. & Image Preprocessing, Edge Detection  &  N/A & N/A  & No \\ 

          & UMI \cite{wickstrom2018uncertainty} & MLSP & CNN+FCN &  SegNet \cite{badrinarayanan2017segnet} & Fully-Sup. & Uncertainty Modeling, Interpretability  &  EndoScene\cite{vazquez2017benchmark} & 547  & No \\ 

          & UNet++ \cite{zhou2018unet} & DLMIA & CNN+UNet &  UNet \cite{ronneberger2015u} & Fully-Sup. & Skip Pathways, Deep Supervision &  ASU-Mayo \cite{tajbakhsh2015automated} & 7,379  & \href{https://github.com/MrGiovanni/UNetPlusPlus}{Open} \\

          & RTPS \cite{wichakam2018real} & MMM & CNN+FCN &  FCN-8S \cite{long2015fully} & Fully-Sup. & Real-time, Compression & EndoScene\cite{vazquez2017benchmark} & 547 & No \\ 

          & MED \cite{nguyen2018colorectal} & AIKE & CNN+FCN &  Deeplabv3 \cite{chen2017rethinking} & Fully-Sup. & Database Augmentation, Multimodal & ClinicDB \cite{bernal2015wm} & 547 & No \\

          & FCNet \cite{brandao2017fully} & SPIE MI &  CNN+FCN & VGG16 \cite{simonyan2015very} & Fully-Sup. & Fully Convolution Networks & ClinicDB \cite{bernal2015wm} & 612 & No \\ 

          & CPFCNet \cite{li2017colorectal} & BMEI &  CNN+FCN & FCN\cite{long2015fully} & Fully-Sup. & Fully Convolution Networks & ClinicDB \cite{bernal2015wm} & 428 & No \\ 

          & APSNet \cite{zhang2017automated} & MIUA & CNN+FCN & FCN-8s \cite{long2015fully} & Fully-Sup. &Region Proposals,  Textons & CVC-ColonDB \cite{tajbakhsh2015automated} & 200  & No \\ 

          & SuperSeg \cite{maghsoudi2017superpixel} & SPMB & N/A  & N/A  & Un-Sup. &Superpixel Segmentation,  SVM & N/A & N/A  & No \\ 

          & APD \cite{yuan2017automatic} & JBHI & MLP & N/A & Un-Sup. & Superpixel, Autoencoder, Saliency & N/A & N/A  & No \\ 

          & WM-DOVA \cite{bernal2015wm} & CMIG & N/A &  N/A & Un-Sup. & Boundary Constraints, Saliency & N/A & N/A  & No \\ 

          & PAD-WCE \cite{jia2015polyps} & ROBIO & N/A &  N/A & Un-Sup. & K-means Clustering, Contour & N/A & N/A  & No \\ 

          & MSA-DOVA \cite{bernal2014polyp} &  TRMI & N/A &  N/A & Un-Sup. & Valley Fnformation, Energy Maps & N/A & N/A  & No \\ 

           \Xhline{0.6pt}
              
\end{tabular}
}}
\caption{Summary of popular deep CPS methods from 2014 to 2020. For the completeness, some early relevant works ranging from 2014 to 2018 have also been stated briefly. }
\label{tab:methods2}
\end{table*} 

\subsubsection{Transformer-based Method}
\label{sec:transformer}
The transformer architecture was first introduced by \cite{vaswani2017attention} published in 2017. Compared to CNN, transformer architecture can capture long-range dependencies in sequences, making it suitable for understanding global context. Recently, Ling \textit{et al.} \cite{ling2023probabilistic} proposed a novel Gaussian-probabilistic guided semantic fusion method for polyp segmentation.
Wang \textit{et al.} \cite{wang2022stepwise} introduce a pyramid transformer architecture for the polyp segmentation task to increase the generalization ability of the neural network. 
Xiao \textit{et al.} \cite{xiao2022icbnet} proposed a transformer-based network for robust and accurate polyp segmentation by mimicking the preliminary-to-refined working paradigm of doctors.
Chen \textit{et al.} \cite{chen2022cld} proposes a novel complement local transformer-based network architecture for medical small object segmentation, which can complement local detailed information when up-sampling global features.
Du \textit{et al.} \cite{du2022swinpa} introduce a novel method called the swin pyramid aggregation network, which introduces two designed modules into the network with a swin transformer as the backbone to learn more powerful and robust features.
Wu \textit{et al.} \cite{wu2022msraformer} presents a multiscale spatial reverse attention network that adopts the Swin Transformer encoder with a pyramid structure to extract the features of four different stages.


\subsubsection{ Hybrid-network Based Method}
\label{sec:hybrid}
Very recently, as shown in Fig. \ref{fig:network_structure}(g), the researcher attempted to combine CNN with transformer for polyp segmentation, aiming at separately capturing local and long-term features from CNN and transformer, respectively.
Jha \textit{et al.} \cite{jha2023transnetr} proposed a transformer-based residual network for colon polyp segmentation and evaluated its diagnostic performance.
Cai \textit{et al.} \cite{cai2022using} proposed a hybrid framework for the polyp segmentation task, in which the encoder consists of a deep Transformer branch and a shallow CNN branch to extract rich features.
Sanderson \textit{et al.} \cite{sanderson2022fcn} proposed a new architecture for polyp segmentation in colonoscopy images, which combines FCNs and transformers to achieve state-of-the-art results.
Huang \textit{et al.} \cite{huang2022transmixer} presented TransMixer, a hybrid interaction fusion architecture of the Transformer branch and the CNN branch, which can enhance the local details of global representations and the global context awareness of local features.
Zhang \textit{et al.} \cite{zhang2022hsnet} introduced a hybrid semantic network by combining CNN and Transformer, aiming at separately capturing local and long-term features in the polyp.
Liu \textit{et al.} \cite{liu2022dbmf} proposed a dual branch multiscale feature fusion network for Polyp Segmentation, which uses CNN and Transformer in parallel to extract multiscale local information and global contextual information, respectively.

\subsection{Level of Supervision}
\label{sec:supervision}
According to the level of supervision, existing deep CPS methods can be categorized into fully-supervised (Sec. \ref{sec:full}), semi-supervised (Sec. \ref{sec:semi}), weakly-supervised (Sec. \ref{sec:weak}), and self-/un-supervised approaches (Sec. \ref{sec:self}).

\subsubsection{Fully-Supervised Methods}
\label{sec:full}
In a fully supervised learning paradigm, a deep model trained on a dataset with pixel-level annotations, which heavily relies on the extensive manual annotation to ensure remarkable performance, and most current deep CPS methods \cite{shen2022task,tomar2022tganet,zhang2022lesion,wang2022stepwise,cai2022using,guo2022non, yang2022source,srivastava2021msrf,sanderson2022fcn,qiu2022bdg,huang2022transmixer,zhao2021automatic} are based on fully supervised learning.
Yin \textit{et al.} \cite{yin2022duplex} proposed a duplex contextual relation network to simultaneously capture the contextual relations across images and within individual images, with 3350 pixel-wise annotations. Zhang \textit{et al.} \cite{zhang2022lesion} designed a lesion-aware dynamic network for the polyp segmentation task with 1348 pixel-wise labels. Wang \textit{et al.} \cite{wang2022stepwise} uses a pyramid Transformer encoder to improve the generalization ability of models. Cai \textit{et al.} \cite{cai2022using} propose a PPFormer for accurate polyp segmentation, which takes advantage of the Transformer’s long-range modeling ability and CNN’s local feature extraction.
Yue \textit{et al.} \cite{yue2022boundary} proposed a novel boundary constraint network where the polyp segmentation task is completed in a fully supervised manner.
Ji \textit{et al.} \cite{ji2022video} design a multistream network for video polyp segmentation, with 19,554 SUN-SEG data for training.

\subsubsection{Semi-Supervised Methods}
\label{sec:semi}
Although a fully-supervised model can achieve high performance, it heavily relies on pixel-wise labeled large-scale datasets. Therefore, more and more researchers are focusing on semi-supervised learning \cite{ijcai2022p155}, which can achieve good performance even with a limited amount of labeled and unlabeled data. For instance, Li \textit{et al.}  proposed a semi-supervised learning paradigm for video polyp segmentation using only 547 pixel-level labeled images, containing a co-training scheme to supervise the predictions of unlabeled images. Zhao \textit{et al.} \cite{zhao2022semi} proposed an accurate and novel network for semi-supervised polyp video segmentation task, which exploits the spatial and temporal information from the proximity frames in endoscope videos
Wu \textit{et al.} \cite{wu2021collaborative} propose a novel semi-supervised polyp segmentation method called collaborative and adversarial learning of focused and dispersive representations learning model.

\subsubsection{Weakly-Supervised Methods}
\label{sec:weak}
On the other hand, to alleviate the fully-supervised methods relying on pixel-wise annotation, some researchers have adopted semi-supervised learning methods, including point supervision, semantic label supervision, scribble supervision, and bounding-box supervision.
Wang \textit{et al.}  \cite{wang2023s2me} proposed a framework of spatial-spectral mutual teaching and ensemble learning for scribble-supervised polyp segmentation.
Wei \textit{et al.} \cite{wei2023weakpolyp} introduced a model completely based on bounding box annotations, reducing the labeling cost and achieving comparable performance to full supervision. 
Wei \textit{et al.} \cite{wei2022boxpolyp} proposed a model to make full use of both accurate mask and extra 40,266 frames box annotation from LDPolypVideo \cite{ma2021ldpolypvideo}.

\subsubsection{Self-/Un-Supervised Methods}
\label{sec:self}

Furthermore, to thoroughly address the dependence on annotated datasets, some researchers have adopted self-supervised methods to pre-train models and transfer them to downstream tasks.
Wang \textit{et al.} \cite{wang2023unsupervised} proposed to use pseudo labels predicted from the source pre-trained model to perform contrastive learning in a supervised way.
Yang \textit{et al.} \cite{yang2021mutual} propose a  progressive self-training module, which selects reliable pseudo labels through a novel uncertainty-guided self-training loss to obtain accurate prototypes in the target domain.
Tian \textit{et al.}  \cite{tian2021constrained} propose a novel self-supervised representation learning method for polyp segmentation, which learns fine-grained feature representations by simultaneously predicting the distribution of augmented data and image contexts using contrastive learning with pretext constraints.
It is worth noting that traditional polyp segmentation methods \cite{sanchez2018automatized,yuan2017automatic,maghsoudi2017superpixel,bernal2015wm,jia2015polyps} are typically implemented in an unsupervised manner. For instance, Jia \textit{et al.}  \cite{jia2015polyps} proposed a feasible method using K-means clustering and localizing region-based active contour segmentation for CPS.


\subsection{Learning Paradigm}
\label{sec:paradigm}
From the perspective of the learning paradigm, deep CPS models can be divided into 12 subcategories, such as attention mechanism, multi-scale feature fusion, boundary awareness, domain adaptation, multi-task learning, adversarial learning, data augmentation, designing loss function, multi-stage refinement, ensemble learning, interpretability, and knowledge distillation.

\subsubsection{Attention Mechanism }

The attention mechanism was first proposed in \cite{bahdanau2014neural} of machine translation, which allows the model to focus on relevant parts of the input when making predictions. Subsequently, it was introduced to the polyp segmentation task.
For instance, Cai \textit{et al.} \cite{cai2022using} present the PP-guided self-attention to enhance the model’s perception of polyp boundary.
Zhao \textit{et al.} \cite{zhao2022semi} propose a novel spatial-temporal attention network composed of temporal local context attention module and proximity frame time-space attention module.
Ji \textit{et al.} \cite{ji2022video} propose a normalized self-attention block, which is motivated by the fact that dynamically updating the receptive field is important for self-attention-based networks.
Huang \textit{et al.} \cite{huang2022transmixer} proposed a hierarchical attention module to encourage the collection of polyp semantic information from high-level features to gradually guide the recovery of polyp spatial information in low-level features.
Patel \textit{et al.} \cite{patel2022fuzzynet} designed a novel attention module to focus more on the difficult pixels that usually lie near the boundary region. 
Du \textit{et al.} \cite{du2022swinpa} proposed a local pyramid attention module to aggregate attention cues in different scales and guide the network to enhance semantic features and emphasize the target area.

\subsubsection{Multiscale Feature Fusion}

The goal of multi-scale feature fusion is to improve the model's ability to deal with varying sizes of objects, leading to more robust and accurate segmentation.
Tomar \textit{et al.} \cite{tomar2022tganet} propose a multi-scale feature aggregation to capture features learned by different decoder blocks.
Wang \textit{et al.} \cite{wang2022stepwise} propose a novel multi-stage feature aggregation decoder, which consists of the local emphasis module and the stepwise feature aggregation module.
Srivastava \textit{et al.} \cite{srivastava2021msrf}  proposed a novel dual-scale dense fusion block that performs dual-scale feature exchange and a sub-network that exchanges multi-scale features.
Yue \textit{et al.} \cite{yue2022boundary} propose a novel cross-layer feature integration strategy that consists of an attention-driven cross-layer feature interaction module and a global feature integration module.
Ji \textit{et al.} \cite{ji2022video} propose a novel global-to-local learning paradigm, which realizes both long-term and short-term spatial-temporal propagation at an arbitrary temporal distance.
Qiu \textit{et al.} \cite{qiu2022bdg} design a boundary distribution guided decoder to fuse multi-scale features to improve polyp segmentation on different sizes.
Huang \textit{et al.} \cite{huang2022transmixer} present the interaction fusion module to bridge the semantic gap between the Transformer branch and the CNN branch, thereby fully capturing the global contextual information and local detailed information of polyps.



\subsubsection{Boundary-aware}

The boundary-aware-based method is designed to address the issue of fuzzy segmentation boundaries in existing polyp segmentation methods.
For example, Jin \textit{et al.} \cite{10113675} introduces an edge extractor to obtain the edge information for auxiliary supervision to use low-level features better. 
Yue \textit{et al.} \cite{yue2022boundary} propose a novel deep network (termed BCNet) by focusing on cross-layer feature integration and boundary extraction in consideration of the challenging characteristics of polyps. With the assistance of high-level location features and boundary constraints, BCNet explores the polyp and
non-polyp information of the shallow layer collaboratively and yields a better segmentation performance.
Qiu \textit{et al.} \cite{qiu2022bdg} design a boundary distribution guided network for accurate polyp segmentation, which consists of a boundary distribution generate module and a boundary distribution guided decoder.
Xiao \textit{et al.} \cite{xiao2022icbnet} proposed an iterative feedback prediction module to consider contextual and boundary-aware information.
Chen \textit{et al.} \cite{chen2022cld} proposes a local edge feature extraction block to progressively generate a sequence of high-resolution local edge features.
Du \textit{et al.} \cite{du2022icgnet} proposed the reverse-contour guidance module to receive contour detail and reverse information to highlight the boundary.


\subsubsection{Domain Adapation}
Domain adaptation addresses the model's performance degradation when there is a distributional shift between the training and testing data.
Wang \textit{et al.} \cite{wang2023unsupervised} a practical problem of source-free domain adaptation, which eliminates the reliance on annotated source data.
Xiong \textit{et al.} \cite{xiong2023unpaired} propose an image-to-image translation network for domain adaptation.
Shen \textit{et al.} \cite{shen2022task} propose a task-relevant feature replenishment-based network to tackle existing problems in unsupervised domain adaptation.
Huy \textit{et al.} \cite{huy2022adversarial} an unsupervised domain adaptation that transfers the style between the domains using Fourier transform
and adversarial training.  
Yang \textit{et al.} \cite{yang2022source} devise a novel source-free domain adaptation framework with Fourier style mining, where only a well-trained source segmentation model is available to adapt to the target domain.
Yang \textit{et al.} \cite{yang2021mutual} propose a mutual-prototype adaptation network to eliminate domain shifts in multi-center and multidevices colonoscopy images.
Li \textit{et al.}  \cite{li2021few} propose a polymorphic transformer, which can be incorporated into any DNN backbones for few-shot domain adaptation.

\begin{table*}[t]
\centering
{
\renewcommand\arraystretch{1.4}
\setlength\tabcolsep{2.5pt}
\resizebox{1\textwidth}{!}{
\begin{tabular}{c|r|r||cccccccc}
\Xhline{1pt} \rule{0pt}{11pt} & \textbf{Datasets} & \textbf{Year} & \textbf{Publication}   & \textbf{Number}    & \textbf{Annotation Type}     & \textbf{Resolution}      & \textbf{Polpy Size ($\%$)} & \textbf{Contrast Values} & \textbf{\#Obj.}& \textbf{Availability } \\\hline\hline
\multirow{10}{*}{\rotatebox{90}{{\textbf{Image-level}}}}

           &ETIS-Larib \cite{silva2014toward} & 2014   & IJCARS   & 196   & Pixel-wise  & 1225$\times$966  & 0.10 $\sim$ 29.1& 0.25 $\sim$ 0.87 & 1 $\sim$ 3&\href{https://drive.google.com/drive/folders/10QXjxBJqCf7PAXqbDvoceWmZ-qF07tFi}{Open} \\

           &CVC-ClinicDB \cite{bernal2015wm} & 2015   & CMIG   & 612   & Pixel-wise  & 384$\times$288  & 0.33 $\sim$ 48.9 & 0.42 $\sim$ 0.96 &1 $\sim$ 3 & \href{https://polyp.grand-challenge.org/CVCClinicDB/}{Open} \\

           &CVC-ColonDB \cite{tajbakhsh2015automated} & 2015   & TMI   & 300   & Pixel-wise  & 574$\times$500  & 0.29 $\sim$ 63.1 & 0.28 $\sim$ 0.93  &1 & \href{https://drive.google.com/drive/folders/1-gZUo1dgsdcWxSdXV9OAPmtGEbwZMfDY}{Open} \\

           &EndoScene \cite{vazquez2017benchmark} & 2017   & JHE   & 912   & Pixel-wise   & \makecell{384$\times$288 to 574$\times$500}& 0.29 $\sim$ 63.2  & 0.27 $\sim$ 0.96  &1 $\sim$ 3 & \href{http://pages.cvc.uab.es/CVC-Colon/}{Open} \\

           &Kvasir-SEG \cite{jha2020kvasir} & 2020   & MMM   & 1,000   & Pixel-wise  & \makecell{332$\times$487 to  1920$\times$1072}  & 0.51 $\sim$ 81.4 & 0.35 $\sim$ 0.87 &1 $\sim$ 3 & \href{https://datasets.simula.no/kvasir-seg/}{Open} \\

           &HyperKvasir \cite{Borgli2020} & 2020   & SD   & 110,079   & \makecell{Pixel-wise, Bounding-box}   & \makecell{332$\times$487 to 1920$\times$1072}  & 0.57 $\sim$ 76.6  & 0.35 $\sim$ 0.88 & 1 $\sim$ 10 & \href{https://osf.io/mh9sj/}{Open} \\

            &Piccolo \cite{sanchez2020piccolo} & 2020   & AS   & 3,433   & Pixel-wise  & \makecell{854$\times$480 to 1920$\times$1080}  & 0.0005 $\sim$ 65.5 &   0.40 $\sim$ 0.72& 1 $\sim$ 5& \href{https://www.biobancovasco.bioef.eus/en/Sample-and-data-catalog/Databases/PD178-PICCOLO-EN.html}{Request} \\

            & Kvasir-sessile \cite{jha2021comprehensive} & 2021 & JBHI & 196 & Pixel-wise &  \makecell{401$\times$415 to 1348$\times$1070}  & 0.54 $\sim$ 58.4 & 0.39 $\sim$ 0.86 & 1 $\sim$ 3 & \href{https://datasets.simula.no/kvasir-seg/}{Open} \\

            & BKAI-IGH \cite{ngoc2021neounet} & 2021 & AVC & 1,200 & \makecell{ Pixel-wise} & \makecell{1280$\times$959}  & 0.14 $\sim$ 19.4 & 0.29 $ \sim$ 0.88 & 1 $\sim$ 18 & \href{https://www.kaggle.com/competitions/bkai-igh-neopolyp/data}{Open} \\

            & PolypGen \cite{ali2023multi} & 2023 & SD & 8,037 & \makecell{Pixel-wise, Bounding-box} & \makecell{384$\times$288 to 1920$\times$1080}  & 0.001 $\sim$ 74.1 & 0.29 $\sim$1.0 & 1 $\sim$ 17 & \href{https://www.kaggle.com/competitions/bkai-igh-neopolyp/data}{Open} \\

           \Xhline{0.6pt}
\multirow{4}{*}{\rotatebox{90}{{\textbf{Video-level}}}}
           & ASU-Mayo \cite{tajbakhsh2015automated} & 2016 & TMI & 36,458 & Pixel-wise & 688$\times$550 & - & - &1 & \href{https://polyp.grand-challenge.org/AsuMayo/}{Request} \\

           & LDPolyVideo \cite{ma2021ldpolypvideo} & 2021 & MICCAI & 901,626 & Bounding-box  & 560$\times$480  & - &0.10 $\sim$ 0.99 &1 & \href{https://github.com/dashishi/LDPolypVideo-Benchmark}{Open} \\

           &ToPV \cite{yao2022scheme} & 2022   & ISBI   & 11,953   &  Video-level   &  384$\times$352 & - &0.08 $\sim$ 1.02 &1 & \href{http://www.pami.sjtu.edu.cn/En/Show/74/163}{Request} \\

           & SUN-SEG \cite{ji2022video} & 2022 & MIR & 158,690 & Diverse-Annotations & 1158$\times$1008 to 1240$\times$1080 & 0.06 $\sim$ 30.0 &0.20 $\sim$ 0.89 &1 &\href{https://github.com/GewelsJI/VPS}{Request} \\

           \Xhline{0.6pt}

\end{tabular}
}}
\caption{Statistics of popular polyp segmentation datasets, including the number of images, annotation type, image resolution, ratio of the object size and contrast value in mages. Please refer to Section. \ref{sec:datasets} for more detailed descriptions.}
\label{tab:datasets}
\end{table*} 

\subsubsection{Multi-task Learning}
Multi-task learning enhances the model's learning capabilities by jointly training on multiple tasks, exploiting complementary information between different tasks to achieve better performance.
Wang \textit{et al.} \cite{wang2023efficient} propose an efficient multi-task synergetic network for concurrent polyp segmentation and classification.
Tomar \textit{et al.} \cite{tomar2022tganet} propose a text-guided attention mechanism for polyp segmentation using a simple byte-pair encoding.
Yue \textit{et al.} \cite{yue2022boundary} proposed a multi-task learning strategy that simultaneously supervises the boundary prediction and polyp prediction with the boundary mask and polyp mask.
Xiao \textit{et al.} \cite{xiao2022icbnet}  presents an iterative context-boundary feedback network to iterative feedback the preliminary predictions
of both segmentation and boundary into different encoder levels.
Murugesan \textit{et al.} \cite{murugesan2019psi} proposed a novel multi-task network and mask prediction is the primary task, while contour detection and distance map estimation are auxiliary tasks.

\subsubsection{Adversarial Learning}

Adversarial learning in image segmentation can enhance models' robustness and generalization capability, making them more suitable for challenging scenarios in the real world.
Mazumdar \textit{et al.}  \cite{10302215} a novel generative adversarial network (GAN) based approach for polyp segmentation, integrating a standard generator and an attention-based discriminator.
Shen \textit{et al.} \cite{shen2022task} propose polyp-aware adversarial learning is employed to bridge the domain gap by aligning features in output space.
Chen \textit{et al.} \cite{chen2022single} designed a random color reversal synthesis module under the GAN framework, which synthesizes images in which some regions are highlighted randomly while the center area of the polyps is always highlighted.
Yang \textit{et al.} \cite{yang2021mutual} propose a novel foreground consistency loss with adversarial learning to optimize the reconstruction module.
Wu \textit{et al.} \cite{wu2021collaborative} proposed auxiliary adversarial learning to improve the quality of segmentation predictions from unlabeled images in the semi-supervised training stage.
Xie \textit{et al.} \cite{xie2020mi} propose a novel MI$^2$GAN to maintain the contents of the medical images during image-to-image domain adaptation.
Poorneshwaran \textit{et al.} \cite{poorneshwaran2019polyp} propose a conditional generative convolutional framework for the task of polyp segmentation.

\subsubsection{Data Augmentation}

Data augmentation enhances the model's generalization and prevents overfitting by applying transformations to the training data, such as rotation, scaling, flipping, etc.
Moreu \textit{et al.} \cite{moreu2023self} an end-to-end model for integrating synthetic and real data under different levels of supervision.
Haithami \textit{et al.} \cite{haithami2023enhancing} propose a deep learning framework to train an image transformation model with a segmentation model jointly.
Guo \textit{et al.} \cite{guo2022non} propose a novel meta-learning mixup data augmentation method and a confidence-aware resampling strategy for polyp segmentation.
Chen \textit{et al.} \cite{chen2022single} presented a random color reversal synthesis module to synthesize polyp images as data augmentation to enhance model performance.
Wei \textit{et al.} \cite{wei2021shallow} propose the color exchange operation to decouple the contents and colors, which reduces the overfitting issue.
Guo \textit{et al.} \cite{guo2020learn} propose a novel ThresholdNet with a confidence-guided manifold mixup data augmentation method to alleviate the limited training dataset and the class imbalance problems.
Thomaz \textit{et al.} \cite{de2019training} proposes a novel method to increase the quantity and variability of training images from a publicly available colonoscopy dataset.

\subsubsection{Designing Loss Function}
In the context of deep CPS, the binary cross-entropy (BCE) is the commonly used loss function. In practice, the BCE loss always fails to learn accurate object boundaries due to the high similarity between polyps and the background tissue. Thus, researchers are dedicated to designing various effective loss functions to address the shortcomings of BCE. 
Wei \textit{et al.}  \cite{wei2023weakpolyp} propose the scale consistency loss to improve the robustness of the model against the variability of the predictions.
Li \textit{et al.} \cite{ijcai2022p155} proposed a context-free loss to mitigate the impact of varying contexts within successive frames.
Wei \textit{et al.} \cite{wei2022boxpolyp}  propose the image consistency loss, which mines supervisory information from the relationship between images.
Zhao \textit{et al.} \cite{zhao2021automatic} build a general training-free loss network to implement the structure supervision in the feature levels, which provides an essential supplement to the loss design based on the prediction itself.
Shen \textit{et al.} \cite{shen2021hrenet} design an edge and structure consistency aware deep loss to improve the segmentation performance further.
Guo \textit{et al.} \cite{guo2020learn} devised a mixup feature map consistency loss and a mixup confidence map consistency loss to exploit the consistent constraints in the training of the augmented mixup data.
Fang \textit{et al.} \cite{fang2020abc} proposed a boundary-sensitive loss designed to leverage the area-boundary constraints.

\subsubsection{Others Techniques}

In addition to the mainstream techniques mentioned above, there are some less commonly used techniques, such as \textit{multi-stage refinement, ensemble learning, interpretability, and knowledge distillation.}
Xiao \textit{et al.} \cite{xiao2022icbnet} propose a multi-stage refinement model iteratively via feedbacking contextual and boundary-aware detail from the preliminary segmentation and boundary predictions. Guo \textit{et al.} \cite{guo2019automated} proposed an ensemble model for automated polyp segmentation by ensembling the UNet, SegNet \cite{badrinarayanan2017segnet}, and PSPNet \cite{zhao2017pyramid}. 
Wickstrom \textit{et al.} \cite{wickstrom2020uncertainty} develop and evaluate recent advances in uncertainty estimation and model interpretability for CPS.
Huang \textit{et al.} \cite{huang2020real} proposed the first knowledge distillation to compress a colonoscopy image segmentation model to achieve real-time segmentation.

\section{Polyp Segmentation Datasets}
\label{sec:datasets}

\begin{figure*}[!h]
\centering


\resizebox{1\textwidth}{!}{
\begin{subfigure}{0.2\textwidth}
\centering
\includegraphics[width=\textwidth]{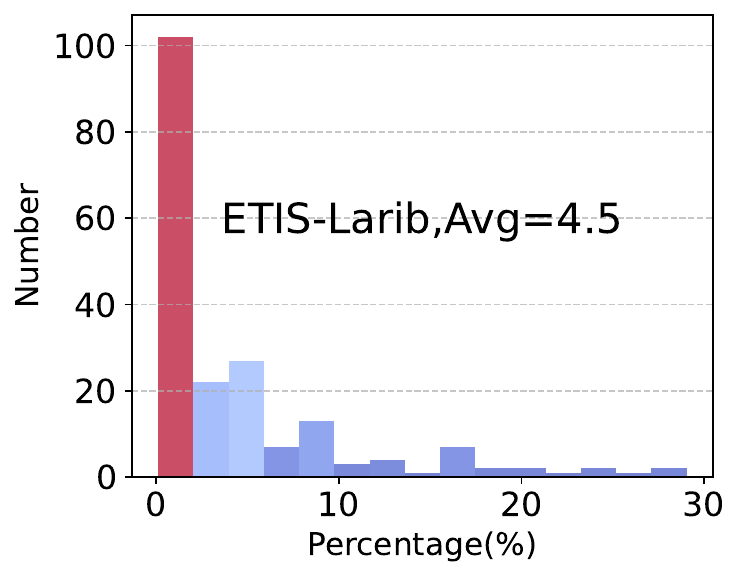}\hspace{-0.8em}
\end{subfigure}
\begin{subfigure}{0.2\textwidth}
\centering
\includegraphics[width=\textwidth]{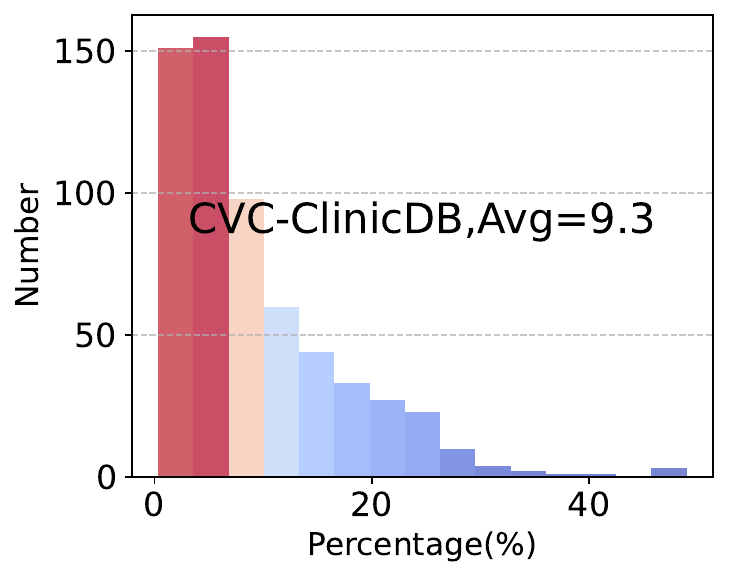}\hspace{-1em}
\end{subfigure}
\begin{subfigure}{0.2\textwidth}
\centering
\includegraphics[width=\textwidth]{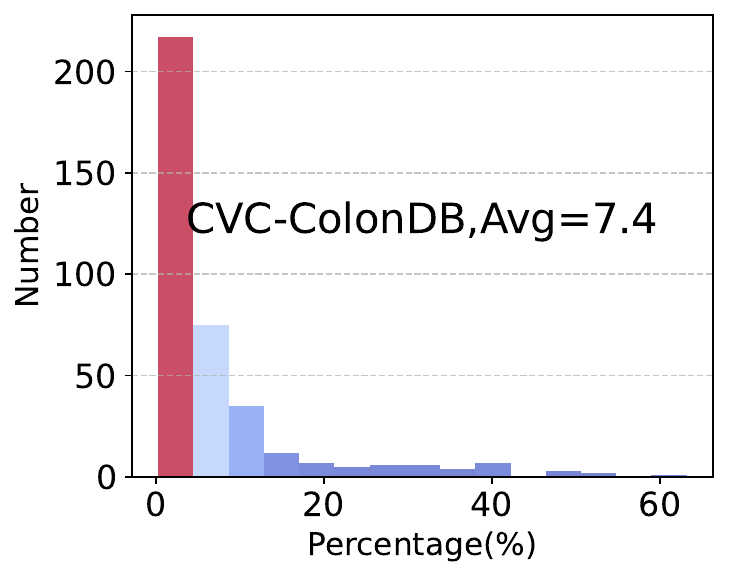}\hspace{-1em}
\end{subfigure}
\begin{subfigure}{0.2\textwidth}
\centering
\includegraphics[width=\textwidth]{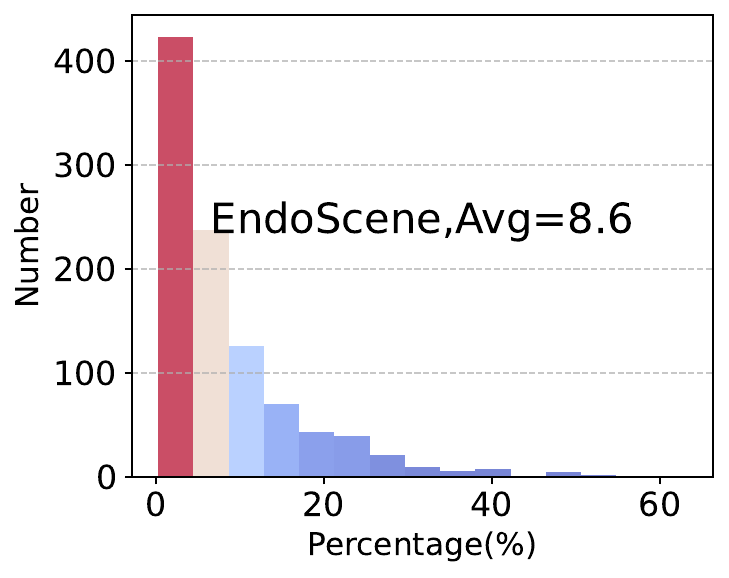}\hspace{-1em}
\end{subfigure}
\begin{subfigure}{0.2\textwidth}
\centering
\includegraphics[width=\textwidth]{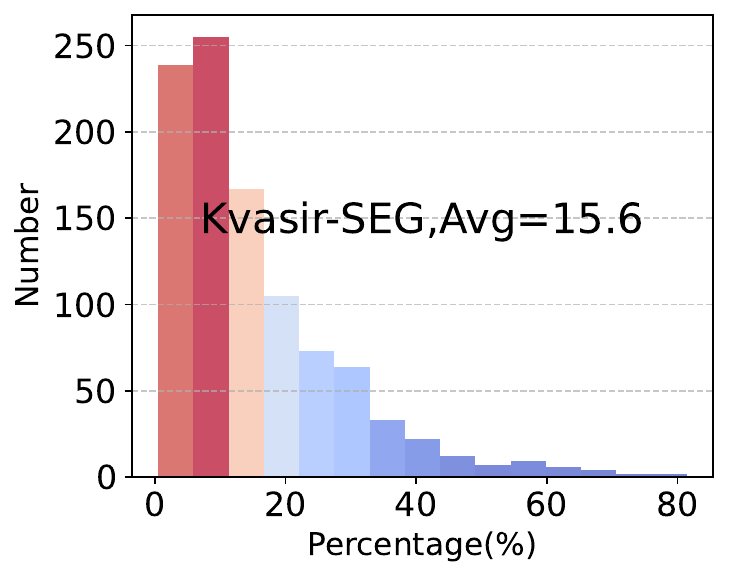}
\end{subfigure}
}

\resizebox{1\textwidth}{!}{
\begin{subfigure}{0.20\textwidth}
\centering
\includegraphics[width=\textwidth]{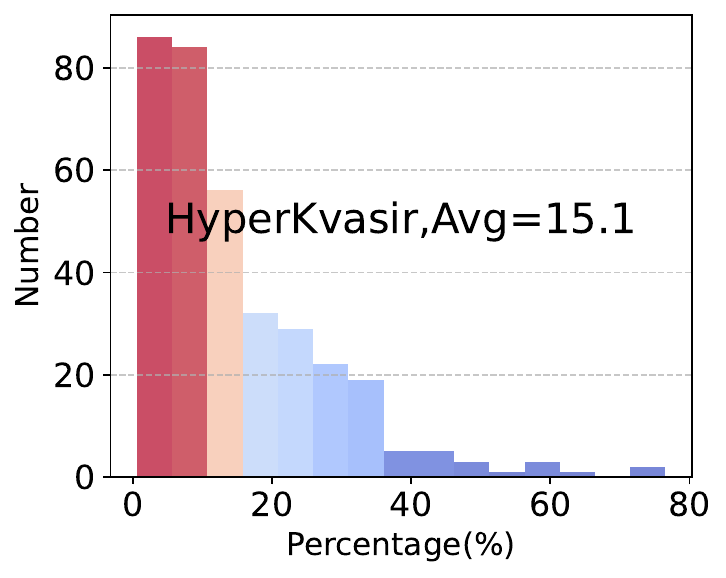}\hspace{-1em}
\end{subfigure}
\begin{subfigure}{0.205\textwidth}
\centering
\includegraphics[width=\textwidth]{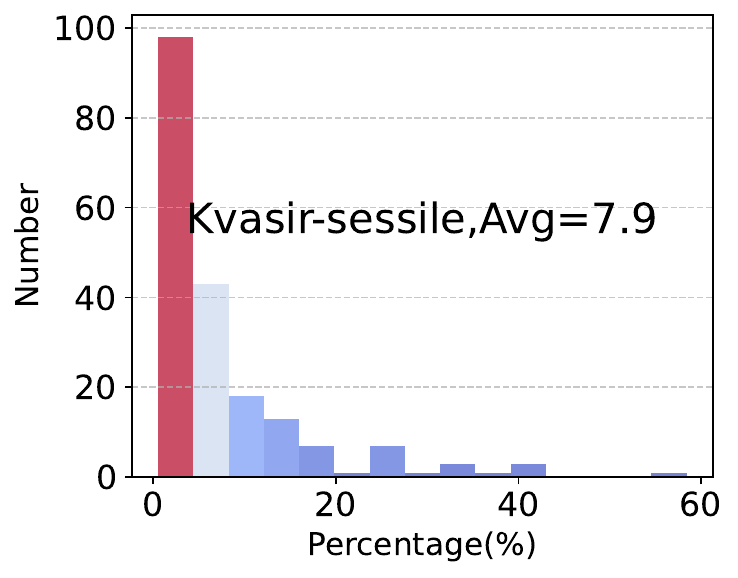}\hspace{-1.3em}
\end{subfigure}
\begin{subfigure}{0.2035\textwidth}
\centering
\includegraphics[width=\textwidth]{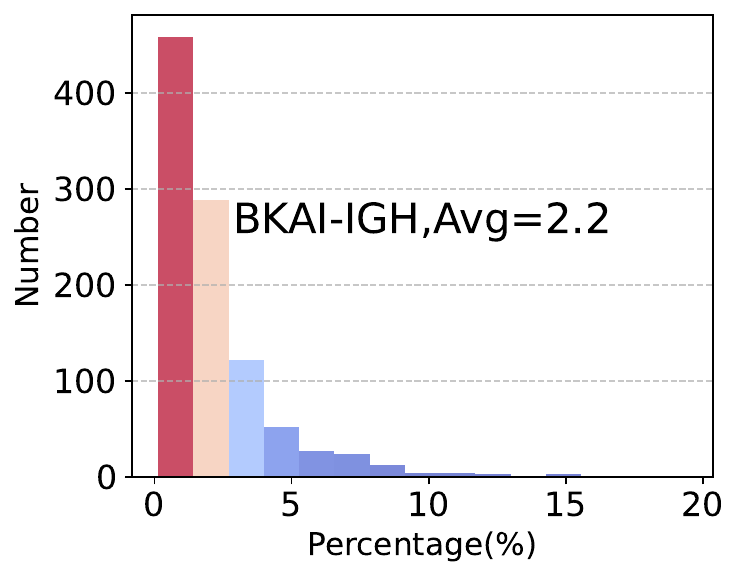}\hspace{-1.5em}
\end{subfigure}
\begin{subfigure}{0.20\textwidth}
\centering
\includegraphics[width=\textwidth]{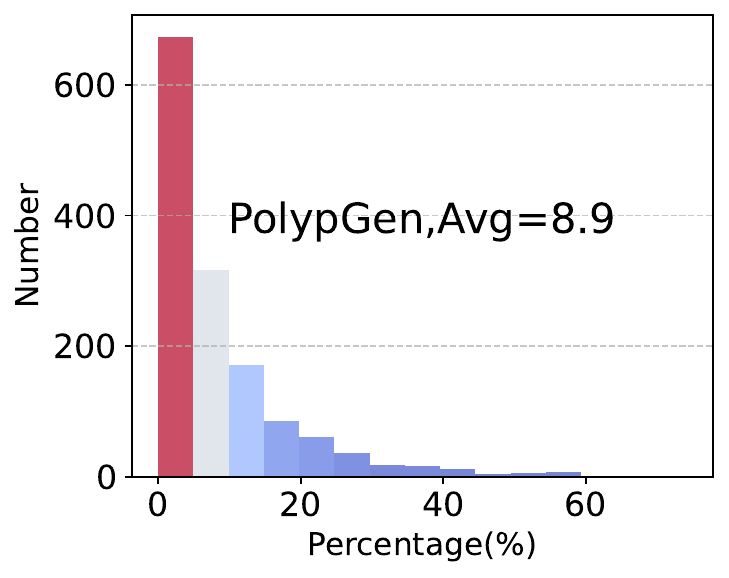}\hspace{-1.0em}
\end{subfigure}
\begin{subfigure}{0.208\textwidth}
\centering
\includegraphics[width=\textwidth]{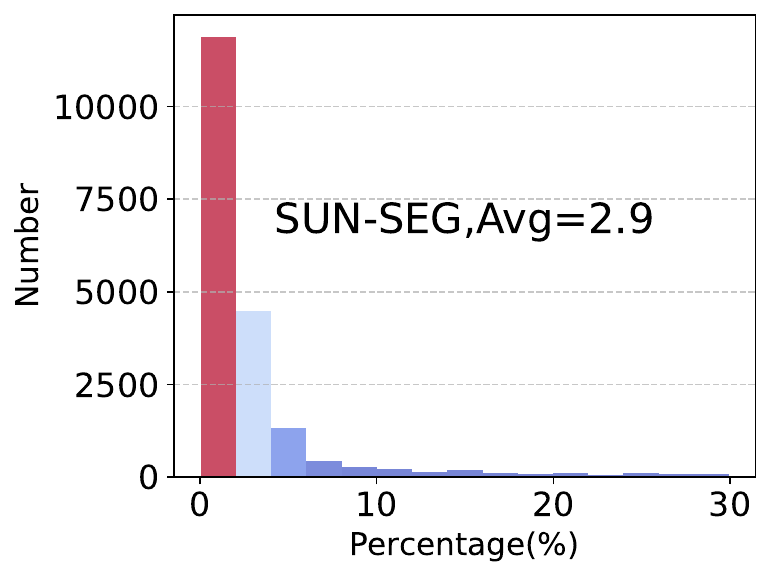}
\end{subfigure}
}


\resizebox{1\textwidth}{!}{
\begin{subfigure}{0.2\textwidth}
\centering
\includegraphics[width=\textwidth]{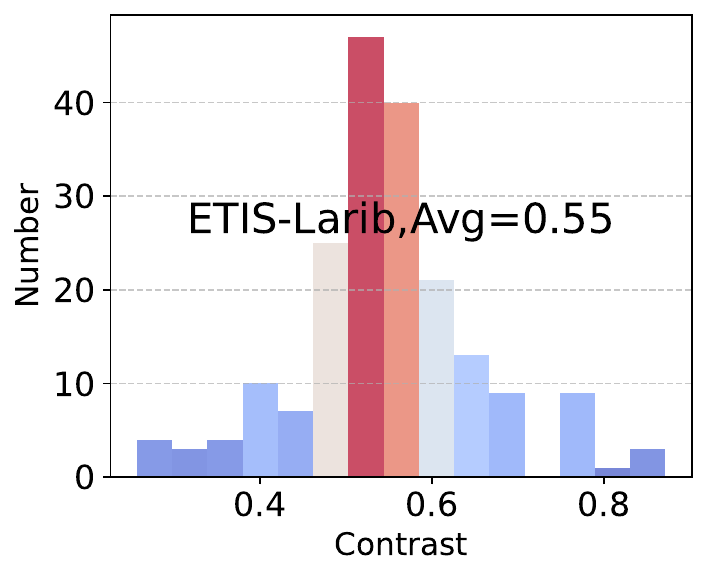}\hspace{-1em}
\end{subfigure}
\begin{subfigure}{0.2\textwidth}
\centering
\includegraphics[width=\textwidth]{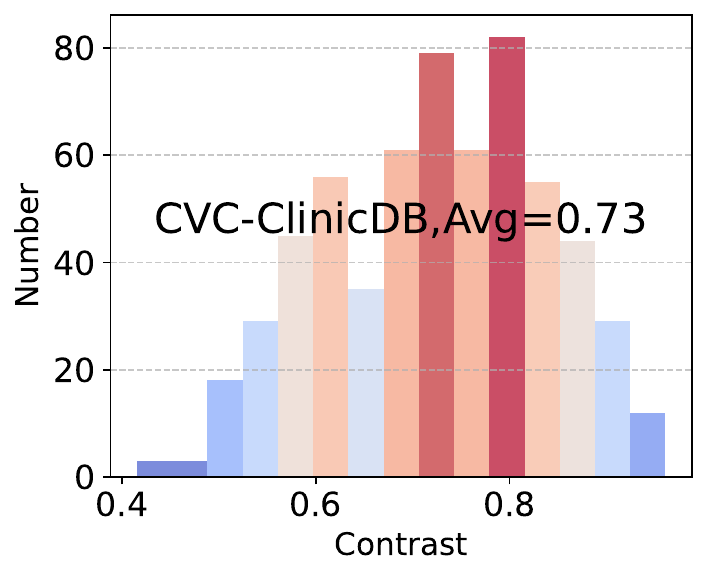}\hspace{-1.5em}
\end{subfigure}
\begin{subfigure}{0.20\textwidth}
\centering
\includegraphics[width=\textwidth]{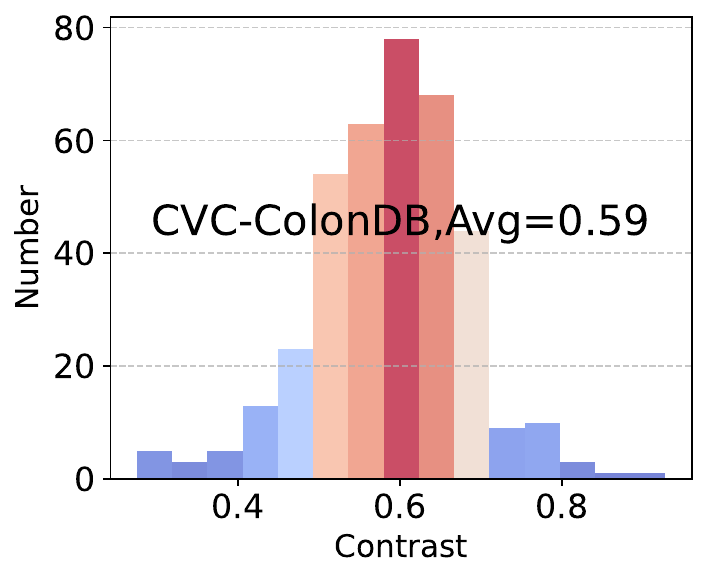}\hspace{-1.8em}
\end{subfigure}
\begin{subfigure}{0.205\textwidth}
\centering
\includegraphics[width=\textwidth]{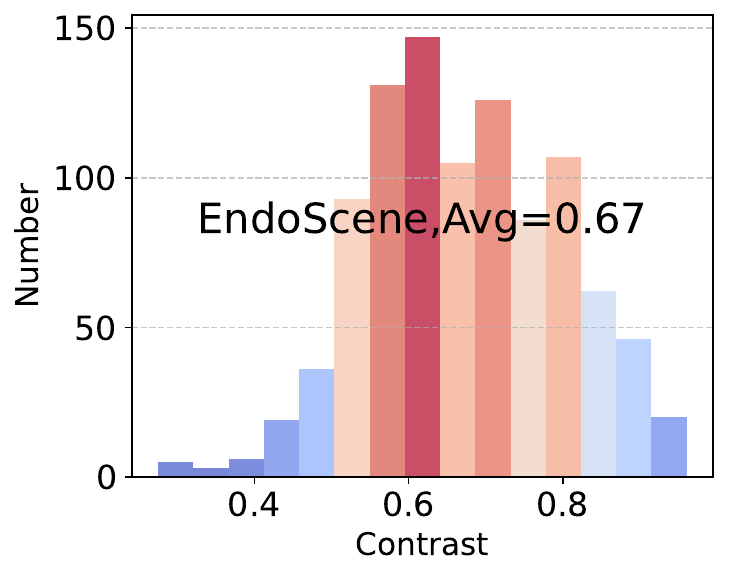}\hspace{-1.5em}
\end{subfigure}
\begin{subfigure}{0.205\textwidth}
\centering
\includegraphics[width=\textwidth]{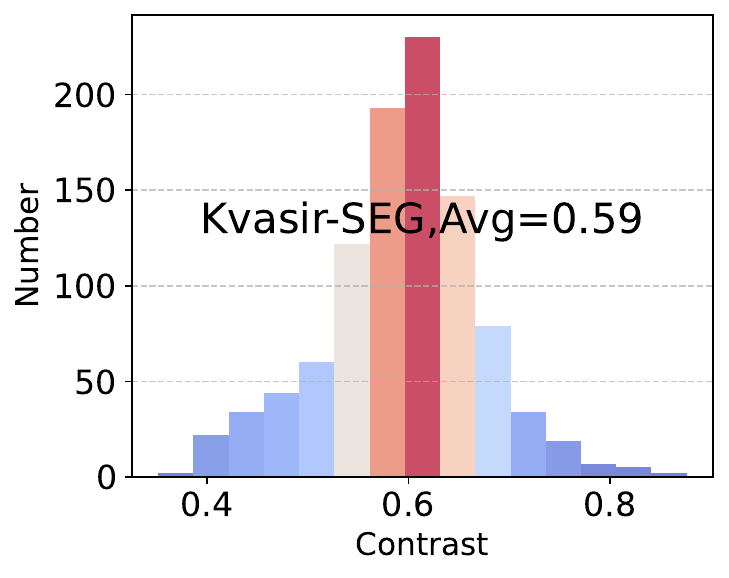}\hspace{-0.51em}
\end{subfigure}
}

\resizebox{1\textwidth}{!}{
\begin{subfigure}{0.205\textwidth}
\centering
\includegraphics[width=\textwidth]{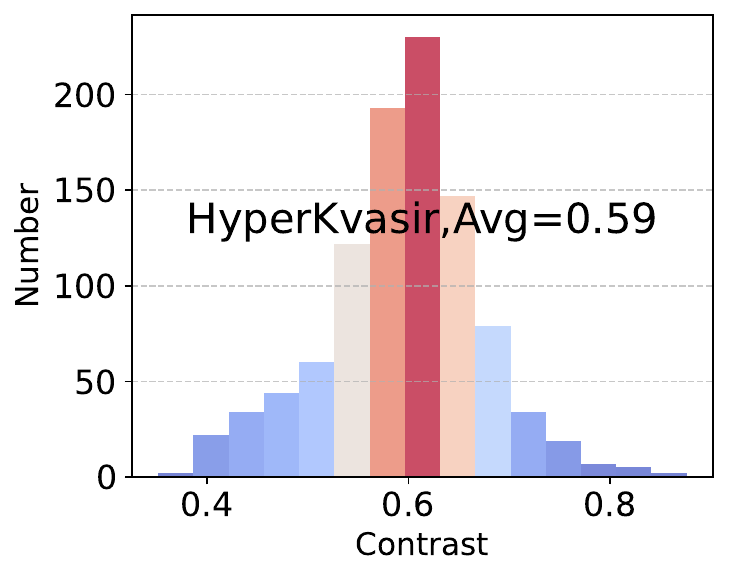}\hspace{-0.5em}
\end{subfigure}
\begin{subfigure}{0.20\textwidth}
\centering
\includegraphics[width=\textwidth]{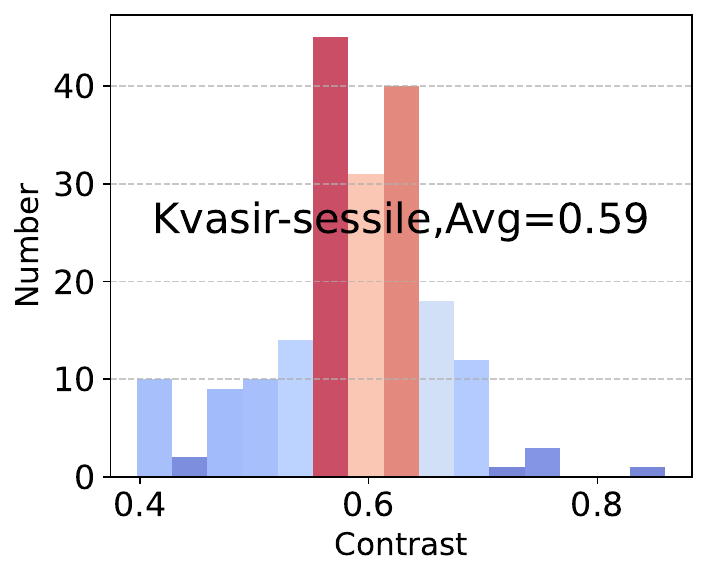}\hspace{-1em}
\end{subfigure}
\begin{subfigure}{0.205\textwidth}
\centering
\includegraphics[width=\textwidth]{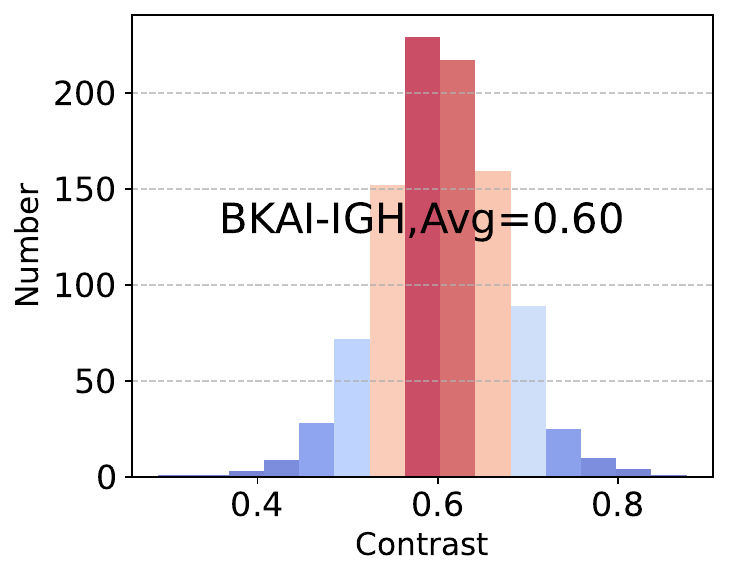}\hspace{-1em}
\end{subfigure}
\begin{subfigure}{0.2\textwidth}
\centering
\includegraphics[width=\textwidth]{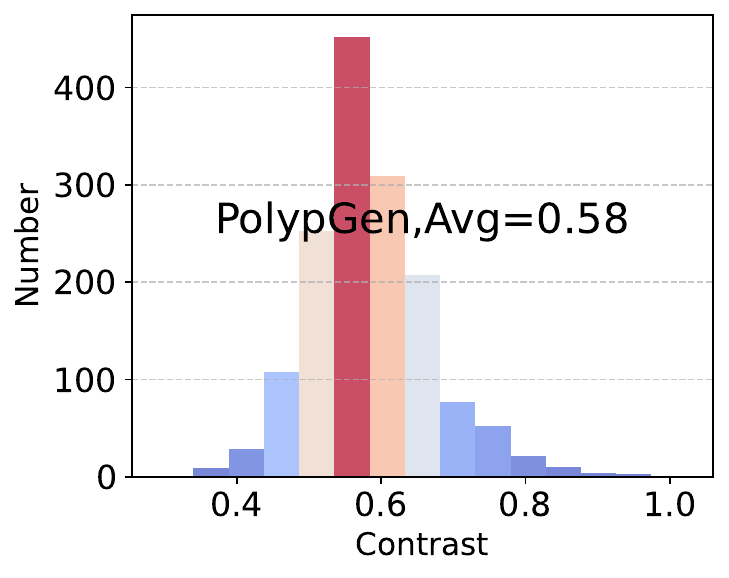}\hspace{-1em}
\end{subfigure}
\begin{subfigure}{0.205\textwidth}
\centering
\includegraphics[width=\textwidth]{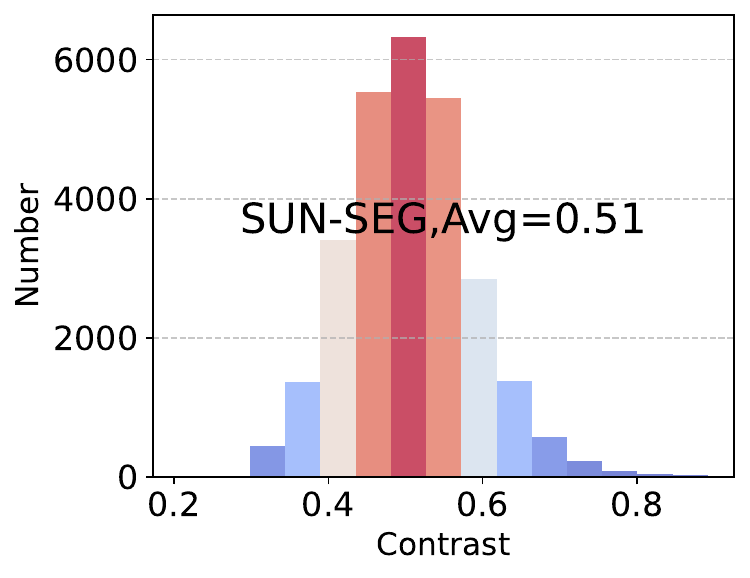}
\end{subfigure}
}

\caption{Statistics of polyp size (first two rows) and contrast value (last two rows) of current CPS datasets. From the perspective of polyp size, the BKAI-IGH is the most challenging dataset because it has the smallest average polyp size, while from the perspective of contrast, the SUN-SEG is the most challenging one because its contrast value is only 0.51.}
\label{fig:contrast_size}
\end{figure*}

\begin{figure}[!t]
\centering
\resizebox{0.485\textwidth}{!}{

\includegraphics[width=\textwidth]{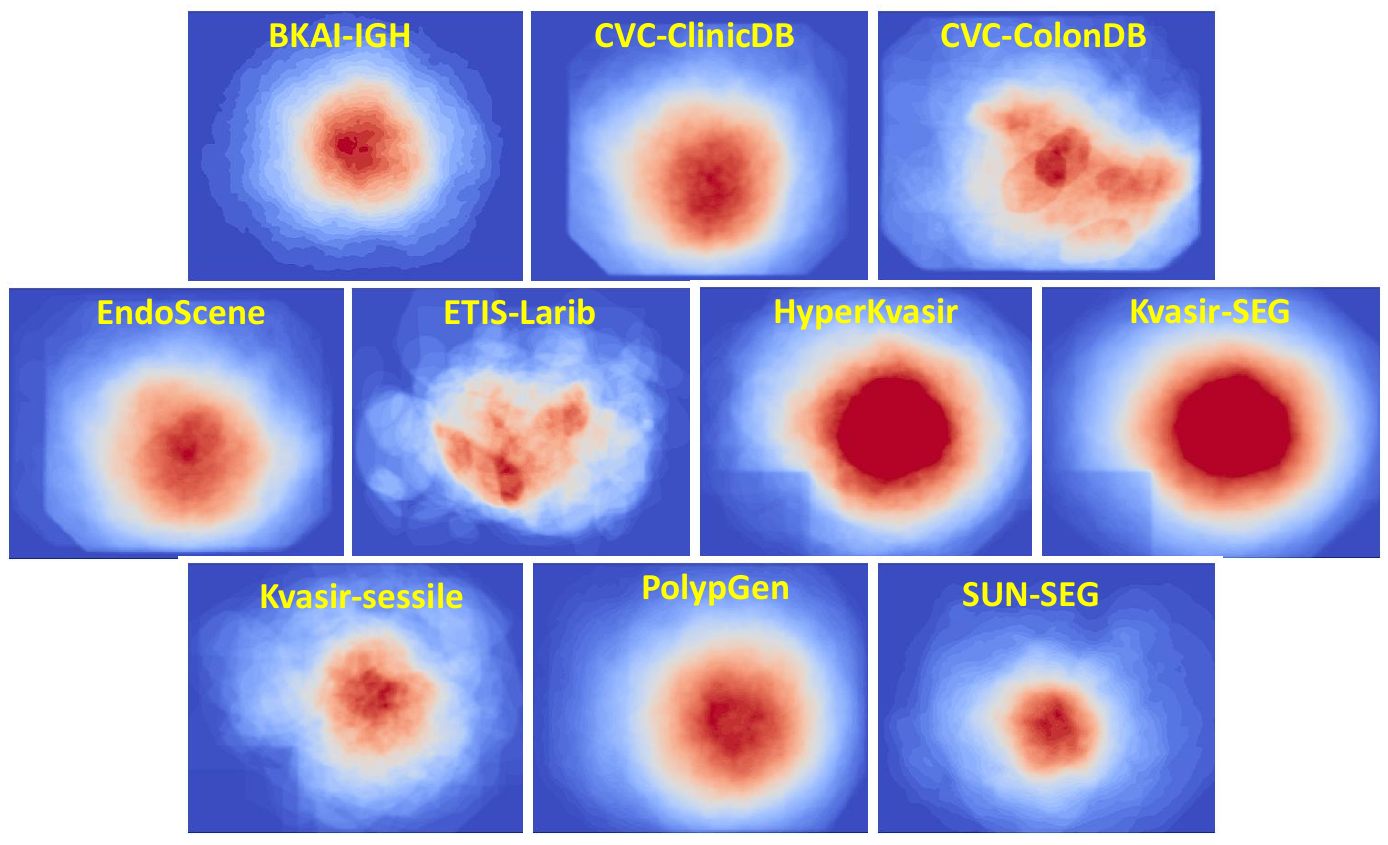}

}

\caption{ Polyps location distributions of some popular CPS datasets. From the perspective of location distribution, the HyperKvasir and Kvasir-SEG datasets are relatively easier. In contrast, the CVC-ColonDB and ETIS-Larib datasets are more challenging because their polyp location distribution is more dispersed.}
\label{fig:heatmap}
\end{figure}

With the rapid development of CPS, many datasets have been introduced. Table \ref{tab:datasets} provides a systematic summary of 14 commonly used CPS datasets, which can be divided into image-level (Sec. \ref{sec:image-level}) and video-level datasets (Sec. \ref{sec:video-level}). 
Besides, as shown in Fig. \ref{fig:contrast_size} and Fig. \ref{fig:heatmap}, we also statistics the polyp size, color contrast, and polyp location distribution, providing a better understanding of the characteristics of existing polyp datasets.

\subsection{Image-level Datasets}
\label{sec:image-level}

\noindent \textbf{ETIS-Larib} \cite{silva2014toward} dataset was introduced in 2014, which contains 196 polyp images and their corresponding pixel-wise groundtruth, with a resolution of 1255$\times$966 pixels. The distribution of image contrast of ETIS-Larib ranges from 0.25 to 0.87, with the majority of images having contrast values between 0.45 and 0.7. The proportion of polyp sizes relative to the entire image varies from 0.1\% to 29\%, with most images having polyp sizes falling between 0.1\% to 10\%. The distribution of polyp locations in the images is scattered.

\noindent \textbf{CVC-ClinicDB} \cite{bernal2015wm} was released in 2015, containing 612 polyp images and the corresponding pixel-wise annotations, with a resolution of 384$\times$288 pixels. The distribution of image contrast of CVC-ClinicDB ranges from 0.42 to 0.96, with most images having contrast values between 0.53 and 0.9, and the average contrast is 0.73. The proportion of polyp sizes relative to the entire image varies from 0.33\% to 48.9\%, with the majority of images having polyp sizes falling between 0.1\% to 10\% and the mean sizes being 9.3\%. The distribution of polyp locations in the images is relatively close to the center of the images.

\noindent \textbf{CVC-ColonDB} \cite{tajbakhsh2015automated} was released in 2015, including 300 polyp images and their corresponding pixel-level labels, with a resolution of 574$\times$500 pixels. The distribution of image contrast of CVC-ColonDB ranges from 0.28 to 0.93, with most images having contrast values between 0.4 and 0.75, and the average contrast is 0.59. The proportion of polyp sizes relative to the entire image varies from 0.29\% to 63.1\%, with the majority of images having polyp sizes falling between 0.3\% to 10\% and the mean sizes being 7.4\%.

\noindent \textbf{EndoScene} \cite{vazquez2017benchmark} combines CVC-ColonDB and CVC-ClinicDB into a new dataset (EndoScene) composed of 912 images obtained from 44 video sequences acquired from 36 patients, with different resolutions ranging from 384$\times$288 to 574$\times$500 pixels. The distribution of image contrast of EndoScene ranges from 0.27 to 0.96, with the majority of images having contrast values between 0.5 and 0.8, and the average contrast is 0.67. The proportion of polyp sizes relative to the entire image varies from 0.29\% to 63.2\%, with most images having polyp sizes falling between 0.3\% to 15\% and the mean sizes being 8.6\%.

\noindent \textbf{Kvasir-SEG} \cite{jha2020kvasir} contains 1000 polyp images and their corresponding groundtruth, with different resolutions ranging from 332$\times$487 to 1920$\times$1072 pixels. The contrast of Kvasir-SEG ranges from 0.35 to 0.87, with most images having contrast values between 0.5 and 0.7, and the average contrast is 0.59. The polyp sizes vary from 0.51\% to 81.4\%, with the majority of images falling between 0.5\% to 20\%, and the mean size being 15.6\%. The distribution of polyp locations in the images is close to the center of the images.

\noindent \textbf{HyperKvasir} \cite{Borgli2020} was collected during real gastro- and colonoscopy examinations at a hospital in Norway and partly labeled by experienced gastrointestinal endoscopists. The HyperKvasir contains 110,079 images but only with 348 pixel-wise annotations. The contrast of HyperKvasir ranges from 0.35 to 0.88, and the average contrast is 0.59. The polyp sizes vary from 0.57\% to 76.6\%, with the majority of images falling between 0.5\% to 16\%, and the mean size is 15.1\%.

\noindent \textbf{Piccolo} \cite{sanchez2020piccolo} contains 3433 manually annotated images (2131 White-Light images and 1302 Narrow-Band images), which is 2203 images for the training set, 897 images for the validation set, and 333 images for the test set, with a resolution of 854$\times$480 to 1920$\times$1080 pixels. Its polyps size ranges from 0.0005\% to 65.5\%, its contrast varies from 0.40 to 0.72, and each image contains 1$\sim$5 polyps.

\noindent \textbf{Kvasir-sessile} \cite{jha2021comprehensive} selected from the Kvasir-SEG \cite{jha2020kvasir}, contains 196 flat or sessile polyps and the corresponding pixel-wise annotations, with a resolution of 401$\times$415 to 1348$\times$1070 pixels. Its polyps size ranges from 0.54\% to 58.4\%, its contrast varies from 0.39 to 0.86, and each image contains 1$\sim$3 polyps.

\noindent \textbf{BKAI-IGH} \cite{ngoc2021neounet} contains 1200 images, with a resolution of 1280$\times$959 pixels. The training set consists of 1000 images with instance-level pixel-wise annotations, and the test set consists of 200 images without annotations. All polyps are also classified into neoplastic or non-neoplastic classes denoted by red and green colors. Its polyps size ranges from 0.14\% to 19.4\%, its contrast varies from 0.29 to 0.88, and each image contains 1$\sim$18 polyps.

\noindent \textbf{PolypGen} \cite{ali2023multi} is composed of a total of 8037 frames, including both single and sequence frames, consisting of 3762 positive sample frames collected from six centers and 4275 negative sample frames collected from four different hospitals. Its polyps size ranges from 0.001\% to 74.1\%, its contrast varies from 0.29 to 1.0, and each image contains 1$\sim$17 polyps.

\subsection{Video-level Datasets}
\label{sec:video-level}

\noindent \textbf{ASU-Mayo} \cite{tajbakhsh2015automated} is composed of a total of 36,458 frames, which is the first and a constantly growing set of short and long colonoscopy videos, collected and de-identified at the Department of Gastroenterology at Mayo Clinic in Arizona. Each frame has a groundtruth or binary mask indicating the polyp region. However, the problem is that it is not publicly available, and as of our submission, the authors have not responded to our requests for the dataset.

\noindent \textbf{ToPV} \cite{yao2022scheme} contains 360 short videos collected during real colonoscopy examination. Two hundred four videos contain one polyp labeled by experienced endoscopic physicians and cropped to a size of 384×352. The maximum duration of video clips in our dataset is 58.4 seconds, and the minimal duration is 5.88 seconds, with a median duration of 15.76 seconds and an average duration of 16.78 seconds. All videos are under white-light endoscopy observation.

\noindent \textbf{LDPolyVideo} \cite{ma2021ldpolypvideo} consists of 160 videos with 40,266 frames and the corresponding bounding box annotations, with a resolution of 560$\times$480. 33,884 frames contain at least one polyp, and in total, 200 labeled polyps. Besides we also provide 103 videos, including 861,400 frames without full annotations. Each video has a label indicating whether it contains polyps.

\noindent \textbf{SUN-SEG} \cite{ji2022video} contains positive cases with 49,136 polyp frames and negative cases with 109,554 non-polyp frames. The SUN-SEG dataset contains diversified annotations, including pixel-wise object mask, boundary, scribble, and polygon annotations. The SUN-SEG dataset consists of 19,544 frames for training and 29,592 frames for testing. The SUN-SEG is the most well-annotated and high-quality dataset for polyp segmentation.

\subsection{Analysis and Discussion}

\begin{figure}[!t]
\centering
\resizebox{0.48\textwidth}{!}{
\includegraphics[width=\textwidth]{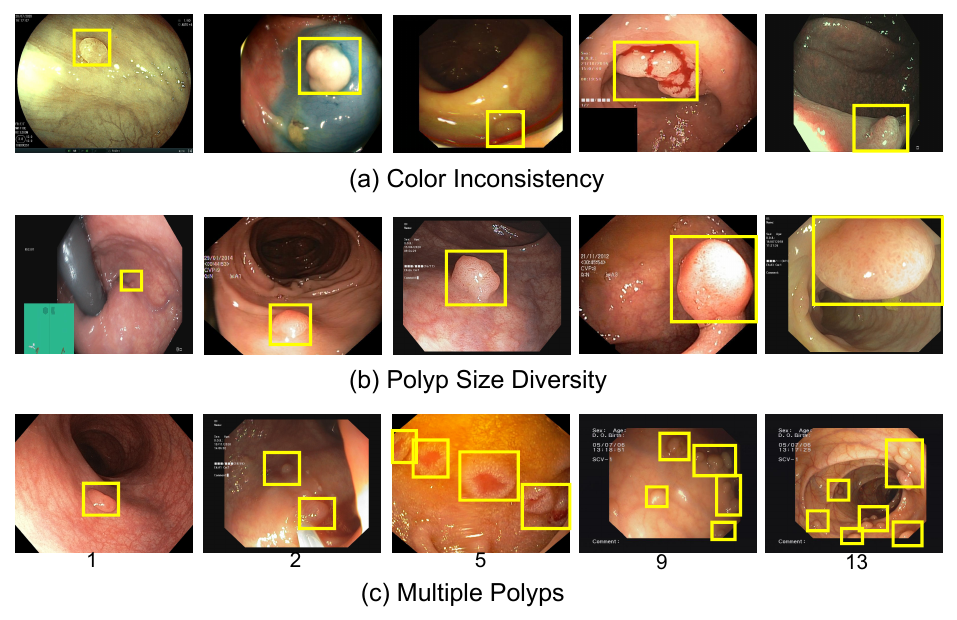}
}
\vspace{-0.5em}
\caption{Illustration of polyp diverse variations in color, size, and quantity, making it very challenging to accurately segment polyps.  }

\label{fig:polyps}
\end{figure}

As shown in Table \ref{tab:datasets}, HyperKvasir is the largest dataset for image polyp segmentation in terms of quantity but only with 348 pixel-wise annotations, while LDPolyVideo is the largest dataset for video polyp segmentation with only bounding box annotations. From the perspective of quality and quantity of annotated data, SUN-SEG is currently the best dataset, with 49,136 well-annotated images, including pixel-wise object mask, boundary, scribble, and polygon annotations. As demonstrated in Fig. \ref{fig:contrast_size}, the BKAI-IGH is the most challenging dataset in terms of polyp size because it has the smallest average polyp size, while from the perspective of contrast, the SUN-SEG is more challenging one because its contrast value is only 0.51. As shown in Fig. \ref{fig:heatmap}, from the perspective of location distribution, the HyperKvasir and Kvasir-SEG datasets are relatively easier, while the CVC-ColonDB and ETIS-Larib datasets are more challenging because their polyp location distribution is more dispersed. Besides, polyps exhibit diverse variations in color, size, and quantity, making it very challenging to accurately segment them, as shown in Fig. \ref{fig:polyps}.


\section{Evaluation Metrics}
\label{sec:metrics}

This section reviews commonly used CPS evaluation metrics, including Dice, IoU, Precision, Recall, Specificity, PR curves, F-measure \cite{achanta2009frequency}, MAE \cite{perazzi2012saliency}, Weighted F-measure \cite{w-fmeasure}, S-measure \cite{Smeasure}, E-measure \cite{fan2018enhanced} and FPS.

\noindent \textbf{Dice coefficient}, also known as the F1 score, is a measure of the overlap between two sets, with a range of 0 to 1. A value of 1 indicates a perfect overlap, while 0 indicates no overlap.

\begin{equation}
\label{distribution difference}
Dice=\frac{2TP}{ 2TP + FP + FN}
\end{equation}

\noindent \textbf{IoU} (Intersection over Union) measures the overlap between two sets but is expressed as a ratio of the size of the intersection to the size of the union of the sets.
\begin{equation}
IoU=\frac{TP}{ TP + FP + FN}
\end{equation}

\noindent \textbf{Precision} is a measure of the positive predictive value of a classifier or the proportion of true positive predictions among all positive predictions.
\begin{equation}
Precision=\frac{TP}{ TP + FP}
\end{equation}

\noindent \textbf{Recall}, also known as sensitivity or true positive rate, is an evaluation metric used in binary and multiclass classification to measure the ability of a model to identify positive instances correctly. It quantifies the proportion of actual positive instances that the model correctly predicted.
\begin{equation}
Recall=\frac{TP}{ TP + FN}
\end{equation}

\noindent \textbf{Accuracy} is the overall correct classification rate or the proportion of correct predictions made by the classifier out of all predictions made.
\begin{equation}
Acc=\frac{TP+TN}{ TP + TN + FP + FN}
\end{equation}

\begin{center}
\begin{table*}[!htb]
\setlength{\abovecaptionskip}{0pt}%
\setlength{\belowcaptionskip}{5pt}%
\small{
\linespread{2}
\renewcommand\arraystretch{1}
\resizebox{0.99\textwidth}{!}{
\begin{tabular}{r|r||cc|cc|cc|cc|cccc}
\Xhline{1pt}
\multirow{2}{*}{Method}
&\multirow{2}{*}{Publication}

& \multicolumn{2}{c|}{CVC-Clinic. \cite{bernal2015wm}}
& \multicolumn{2}{c|}{Kvasir-SEG \cite{jha2020kvasir}}
& \multicolumn{2}{c|}{CVC-ColonDB \cite{tajbakhsh2015automated}}
& \multicolumn{2}{c|}{ETIS-Larib \cite{silva2014toward}}
& \multicolumn{2}{c|}{EndoScene \cite{vazquez2017benchmark}}

\\
\cline{3-12}
&   & mDice & mIoU & mDice & mIoU & mDice & mIoU & mDice & mIoU & mDice & mIoU \\
\hline\hline
RPFA \cite{su2023revisiting}    &MICCAI 2023        & 0.931 &0.885    &\cellcolor{B}{\textbf{\textcolor{white}{0.929}}} &\cellcolor{R}{\textbf{\textcolor{white}{0.880}}}   & \cellcolor{R}{\textbf{\textcolor{white}{0.837}}}&\cellcolor{R}{\textbf{\textcolor{white}{0.759}}}      & \cellcolor{R}{\textbf{\textcolor{white}{0.822}}} & \cellcolor{R}{\textbf{\textcolor{white}{0.746}}}        &0.905 &0.839     \\

XBFormer \cite{wang2023xbound} & TMI 2023         & 0.923 & 0.875    & 0.926 & 0.871   &0.808 & 0.724     & 0.738 & 0.650       & 0.868 & 0.791  \\

ColnNet \cite{jain2023coinnet} & TMI 2023         & 0.930 & 0.887    & 0.926 & 0.872   & 0.797 & 0.729    & 0.759 & 0.690       & 0.909 &0.863  \\

FSFM \cite{su2023accurate}  & ISBI 2023           & 0.934 & 0.884    & 0.913 & 0.861   & 0.786 & 0.709    & 0.778 & 0.702       & \cellcolor{B}{\textbf{\textcolor{white}{0.910}}} & \cellcolor{B}{\textbf{\textcolor{white}{0.846}}}  \\

CASCADE \cite{rahman2023medical} & WACV 2023     & 0.943  & 0.899    & 0.926  & 0.878  & 0.825  & 0.745   & \cellcolor{B}{\textbf{\textcolor{white}{0.801}}} & \cellcolor{B}{\textbf{\textcolor{white}{0.728}}}       & 0.905 & 0.838 \\

PolypPVT \cite{dong2023PolypPVT} & CAAI AIR 2023   &0.937 &0.889     &0.917 &0.864     &0.808 &0.727      &0.787 & 0.706         &0.900 & 0.833  \\

RealSeg \cite{su2023go} & ISBI 2023               & 0.923 & 0.873    & 0.913 & 0.863   & 0.785 & 0.710    & 0.777 & 0.698       &0.909 &0.844    \\

Polyp-Mixer \cite{shi2022polyp} & TCSVT 2023      & 0.908 & 0.856    & 0.916 & 0.864   & 0.791 & 0.706    & 0.759 & 0.676       & -     & -   \\

CFANet \cite{zhou2023cross} & PR 2023            & 0.933  & 0.883    & 0.915 & 0.861   & 0.743 & 0.665    & 0.732 & 0.655       & 0.893 & 0.827 \\

\textbf{M2SNet} \cite{zhao2023m} & Arxiv 2023     & 0.922 & 0.880    & 0.912 & 0.861   & 0.758 & 0.685    & 0.749 & 0.678       & 0.869 & 0.807 \\

EMTSNet \cite{wang2023efficient} & JBHI 2023      &  0.935 & 0.885   & 0.919 &0.869    & 0.788 & 0.708    & 0.780 & 0.702       & 0.900 & 0.833  \\ \hline

PPFormer \cite{cai2022using} & MICCAI 2022       & \cellcolor{R}{\textbf{\textcolor{white}{0.946}}}  &\cellcolor{R}{\textbf{\textcolor{white}{0.902}}}     & \cellcolor{R}{\textbf{\textcolor{white}{0.930}}} & \cellcolor{B}{\textbf{\textcolor{white}{0.879}}}   & \cellcolor{B}{\textbf{\textcolor{white}{0.823}}} & \cellcolor{B}{\textbf{\textcolor{white}{0.756}}}    & 0.791 & 0.706       & \cellcolor{R}{\textbf{\textcolor{white}{0.919}}} & \cellcolor{R}{\textbf{\textcolor{white}{0.857}}}  \\

SSFormer-L \cite{wang2022stepwise} & MICCAI 2022  & 0.906 & 0.855   &0.917 &0.864      & 0.802 &0.721      &0.796 & 0.720        & 0.895 &0.827  \\

SSFormer-S \cite{wang2022stepwise} & MICCAI 2022  &0.916 & 0.873      &0.925 &0.878    &0.772 & 0.697      &0.767 & 0.698        &0.887 & 0.821 \\

LDNet \cite{zhang2022lesion} & MICCAI 2022        & 0.923 & 0.872      &0.912 & 0.855   & 0.794 & 0.715    & 0.778 & 0.707       &0.893 & 0.826   \\

TGANet \cite{tomar2022tganet} & MICCAI 2022     & 0.863 & 0.805      & 0.886 & 0.822   & 0.695 & 0.609     & 0.574 & 0.488       & 0.822 & 0.733 \\

TransMixer \cite{huang2022transmixer} & BIBM 2022 & \cellcolor{B}{\textbf{\textcolor{white}{0.945}}}& \cellcolor{B}{\textbf{\textcolor{white}{0.900}}}    & 0.923 & 0.876    &0.823 & 0.745     &  0.795 & 0.719      & \cellcolor{B}{\textbf{\textcolor{white}{0.910}}} & 0.844 \\

ICBNet \cite{xiao2022icbnet} & BIBM 2022         & 0.938 & 0.892    & 0.928& 0.883     &0.812 & 0.738      &0.800&  0.727       & 0.898 & 0.833  \\

TASNet \cite{chen2022single} & BIBM 2022       & 0.930 & 0.884       & 0.913 & 0.863   &0.799   & 0.719   & 0.797 & 0.720      & 0.894  & 0.825  \\

DCRNet \cite{yin2022duplex}  & ISBI 2022         & 0.896 & 0.844     & 0.886 & 0.825   & 0.704 & 0.631     & 0.556 & 0.496       & 0.856 & 0.788  \\

BDGNet \cite{qiu2022bdg} & SPIE MI 2022      &  0.905 & 0.857     & 0.915 & 0.863   & 0.797 & 0.723     &  0.752 & 0.681      & 0.899 & 0.831   \\

Conv-MLP \cite{jin2023polyp} & VisCom 2022      & 0.924 & 0.870      & 0.920 & 0.869   & 0.793 & 0.717     & 0.753 &  0.676      & 0.893 & 0.822 \\

GLFRNet \cite{song2022global} & TMI 2022 &     0.941 & 0.895        & 0.894 & 0.837     & 0.729 & 0.659     &0.674 & 0.595       & 0.898 & 0.827  \\

FANet \cite{tomar2022fanet} & TNNLS 2022    & 0.823 &0.756         & 0.852 & 0.791     & 0.558 & 0.486     &0.415 & 0.361       &0.668 & 0.600   \\

FNet-Res2Net \cite{patel2022fuzzynet} & NeurIPS 2022 &  0.919 & 0.867&  0.889       & 0.830 & 0.739     &0.662 & 0.731      &0.658 &0.894 &0.825 \\

FNet-PVT \cite{patel2022fuzzynet} & NeurIPS 2022 & 0.937 & 0.889 & 0.913 & 0.864       &0.811 & 0.728    & 0.791 & 0.702      & 0.891  & 0.818  \\

\textbf{CaraNet}\cite{lou2022caranet} &SPIE MI 2022& 0.921 & 0.876     & 0.913 & 0.859   & 0.775 & 0.700    & 0.740 & 0.660      & 0.902 & 0.836 \\ \hline

Transfuse \cite{zhang2021transfuse} & MICCAI 2021   & 0.908 & 0.857   & 0.915 & 0.860    & 0.790 & 0.710    & 0.748 & 0.657      & 0.893 & 0.825 \\

MSNet \cite{zhao2021automatic} & MICCAI 2021    & 0.915 &0.866      &0.902  &0.847      &0.747 &0.668      &0.720 & 0.650        &0.862  &0.796 \\

SANet \cite{wei2021shallow} & MICCAI 2021       &0.916 &0.859      &0.904 & 0.847      &0.752 & 0.669      &0.750 & 0.654        &0.888  & 0.815 \\


UACANet-L \cite{kim2021uacanet} & ACM MM 2021      & 0.926 & 0.880      & 0.912 & 0.859    &0.751 & 0.678    &0.766   &0.689       &0.909   &0.844 \\

UACANet-S \cite{kim2021uacanet} & ACM MM 2021    & 0.916 & 0.870      &0.905 & 0.852     &0.783 & 0.704     &0.694  & 0.615        & 0.902   &0.837 \\


EUNet \cite{patel2021enhanced} & CVR 2021      & 0.902 & 0.846    & 0.908 & 0.854     & 0.756 & 0.681      & 0.687 & 0.609       & 0.837 & 0.765 \\

EMSNet \cite{wang2021ems} & EMBC 2021         & 0.923  & 0.874     & 0.897 &  0.842   & 0.715 & 0.642      &  0.682 & 0.611      & 0.900 & 0.834  \\

\textbf{MSEG}\cite{huang2021hardnet} & Arxiv 2021 &0.909 &0.864   & 0.897 & 0.839     & 0.735 & 0.666      & 0.700 & 0.630       & 0.874 & 0.804  \\ 

PraNet \cite{fan2020pranet} & MICCAI 2020      & 0.899 & 0.849     &0.898 & 0.840      &0.709 & 0.640      &0.628 & 0.567        &0.871  &0.797  \\

ACSNet \cite{zhang2020adaptive} & MICCAI 2020  & 0.882 & 0.826     & 0.898 & 0.838     & 0.716 & 0.649     & 0.578 & 0.509       & 0.863 & 0.787 \\ 

SFANet \cite{fang2019selective} & MICCAI 2019  & 0.700 & 0.607    & 0.723 & 0.611     & 0.469 & 0.347      &0.297 & 0.217       & 0.467 & 0.329 \\ 

ResUNet++ \cite{jha2019resunet} & ISM 2019     & 0.846 &0.786      &0.807 & 0.727     & 0.588 & 0.497      &0.337 & 0.275        &0.687 & 0.598 \\

UNet \cite{ronneberger2015u} & MICCAI 2015    & 0.823 & 0.755     &0.818 & 0.746     & 0.512 & 0.444       &0.398 & 0.335       & 0.710 & 0.627 \\

\Xhline{1pt}
 \end{tabular}  }}
\vspace{0.5em}
\caption{Benchmarking performance of 40 state-of-the-art deep CPS models on five commonly used datasets in terms of mDice and mIoU. The top 2 methods are marked in \textbf{\textcolor{R}{Red}} and \textbf{\textcolor{B}{Blue}}, respectively.}
\label{tab:results}
\end{table*}
\end{center}


\noindent \textbf{Specificity}, also known as True Negative Rate, is an evaluation metric used in binary and multiclass classification to measure the ability of a model to identify negative instances correctly. 
\begin{equation}
Specificity=\frac{TN}{  TN + FP }
\end{equation}

\noindent  {\textbf{PR curves}} provides a visual representation of the performance of a classification model across various levels of precision and recall by plotting precision against recall at different classification thresholds.

\noindent {\textbf{F-measure}} \cite{achanta2009frequency} is a harmonic mean of average precision and average recall. We compute the F-measure as
\begin{equation}
F_{\beta}=\frac{(1 + \beta^2) \times {\rm Precision} \times {\rm Recall}}{\beta^2 \times {\rm Precision + Recall}} 
\end{equation}
where we set $\beta^2$ to be 0.3 to weigh precision more than recall.

\noindent  {\textbf{MAE}} (Mean Absolute Error) \cite{perazzi2012saliency} is calculated as the average pixel-wise absolute difference between the binary ${GT}$ and the saliency map ${S}$.
\begin{equation}
\label{MAE}
MAE=\frac{1}{ W \times H} \sum \limits_{x=1}^{W} \sum \limits_{y=1}^{H} \Big|{S}(x,y) - {GT}(x,y)\Big| 
\end{equation}
where $W$ and $H$ are width and height of the saliency map $S$, respectively.

\noindent  {\textbf{Weighted F-measure}}. Weighted F-measure \cite{w-fmeasure}  define weighted precision, which is a measure of exactness, and weighted recall, which is a measure of completeness:
\begin{equation}
F_{\beta}^w=\frac{(1 + \beta^2) \times {\rm Precision}^w \times {\rm Recall}^w }{ \beta^2 \times {\rm Precision}^w  + {\rm Recall}^w}
\end{equation}

\noindent  {\textbf{S-measure}} ~\cite{Smeasure} simultaneously evaluates region-aware $S_r$ and object-aware $S_o$ structural similarity between the saliency map and ground truth. It can be written as follows: 
\begin{equation}
S_m = \alpha \times S_o + (1 - \alpha) \times S_r
\end{equation}
where $\alpha$ is set to 0.5.

 \noindent  \textbf{E-measure} \cite{fan2018enhanced} combines local pixel values with the image-level mean value to jointly evaluate the similarity between the prediction and the ground truth.

 \noindent \textbf{FPS} (Frame Per Second) is commonly used to assess the efficiency of model inference. FPS is a crucial indicator for evaluating a model's inference speed and real-time performance.

\begin{center}
\begin{table}[t]
\setlength{\abovecaptionskip}{0pt}%
\setlength{\belowcaptionskip}{5pt}%
\LARGE{
\linespread{2}
\renewcommand\arraystretch{1}
\resizebox{0.5\textwidth}{!}{
\begin{tabular}{r||cccc||cccc}
\Xhline{1pt}
\multirow{2}{*}{Method}

& \multicolumn{4}{c||}{\textbf{PolypGen-C1}}
& \multicolumn{4}{c}{\textbf{PolypGen-C2}}
\\
\cline{2-9}
  &  mIoU & mDice & Rec. & Prec. & mIoU & mDice & Rec. & Prec.  \\
\hline\hline
       
UNet \cite{ronneberger2015u}        &.577 &.647 &.678 &.846      &.570 &.634 &.735 &.737   \\

UNet++ \cite{zhou2019unet++}       &.586 &.661 &.695 &.825  &.561 &.624 &.719 &\cellcolor{B}{\textbf{\textcolor{white}{.763}}}    \\

ResUNet++ \cite{jha2019resunet}       &.420 &.524 &.639 &.579  &.278 &.343 &.500 &.420  \\

MSEG \cite{huang2021hardnet}       &.626 &.712  &\cellcolor{R}{\textbf{\textcolor{white}{.780}}} &.793  &.567 &.631 &.727 &.715 \\


LDNet \cite{zhang2022lesion}      &\cellcolor{B}{\textbf{\textcolor{white}{.639}}} &\cellcolor{B}{\textbf{\textcolor{white}{.719}}} &\cellcolor{B}{\textbf{\textcolor{white}{.755}}} &\cellcolor{B}{\textbf{\textcolor{white}{.848}}}   &\cellcolor{B}{\textbf{\textcolor{white}{.609}}} &\cellcolor{B}{\textbf{\textcolor{white}{.689}}} &\cellcolor{R}{\textbf{\textcolor{white}{.854}}} &.687 \\

 TGANet \cite{tomar2022tganet} &.448 &.539 &.642 &.691  &.378 &.458 &.637 &.524 \\

XBFormer \cite{wang2023xbound}       &\cellcolor{R}{\textbf{\textcolor{white}{.654}}} &\cellcolor{R}{\textbf{\textcolor{white}{.720}}} &.744 &\cellcolor{R}{\textbf{\textcolor{white}{.878}}} &\cellcolor{R}{\textbf{\textcolor{white}{.661}}} &\cellcolor{R}{\textbf{\textcolor{white}{.723}}} &\cellcolor{B}{\textbf{\textcolor{white}{.807}}} &\cellcolor{R}{\textbf{\textcolor{white}{.810}}} \\

\Xhline{1pt}


\multirow{1}{*}{}
& \multicolumn{4}{c||}{\textbf{PolypGen-C3}}
& \multicolumn{4}{c}{\textbf{PolypGen-C4}}\\ 

\hline

 UNet \cite{ronneberger2015u}  &.677 &\cellcolor{B}{\textbf{\textcolor{white}{.748}}} &.764 &.879  &.370 &.415 &.655 &.598 \\

 UNet++ \cite{zhou2019unet++} &.653 &.725 &.753 &.857  &.381 &.420 &.634 &.610 \\

 ResUNet++ \cite{jha2019resunet}  &.410 &.511 &.646 &.548  &.169  &.227 &.634 &.282 \\

 MSEG \cite{huang2021hardnet}   & .662 & .744 & \cellcolor{R}{\textbf{\textcolor{white}{.795}}} & .818  & .352 & .394 & .676 & .553  \\


LDNet \cite{zhang2022lesion}    & \cellcolor{B}{\textbf{\textcolor{white}{.707}}} & \cellcolor{R}{\textbf{\textcolor{white}{.787}}} & \cellcolor{R}{\textbf{\textcolor{white}{.795}}} & \cellcolor{B}{\textbf{\textcolor{white}{.889}}} & \cellcolor{B}{\textbf{\textcolor{white}{.427}}} & \cellcolor{B}{\textbf{\textcolor{white}{.483}}} & \cellcolor{R}{\textbf{\textcolor{white}{.737}}} & \cellcolor{B}{\textbf{\textcolor{white}{.630}}}  \\

 TGANet \cite{tomar2022tganet} & .465 & .553 & .626 & .687 & .226 & .276 & .665 & .355 \\

XBFormer \cite{wang2023xbound}   & \cellcolor{R}{\textbf{\textcolor{white}{.722}}} & \cellcolor{R}{\textbf{\textcolor{white}{.787}}} & \cellcolor{B}{\textbf{\textcolor{white}{.790}}} & \cellcolor{R}{\textbf{\textcolor{white}{.913}}} & \cellcolor{R}{\textbf{\textcolor{white}{.460}}} & \cellcolor{R}{\textbf{\textcolor{white}{.504}}} & \cellcolor{B}{\textbf{\textcolor{white}{.687}}} & \cellcolor{R}{\textbf{\textcolor{white}{.714}}} \\

\Xhline{1pt}

\multirow{1}{*}{}
& \multicolumn{4}{c||}{\textbf{PolypGen-C5}}
& \multicolumn{4}{c}{\textbf{PolypGen-C6}}\\ 
\hline

UNet \cite{ronneberger2015u}  &.296 &.361 &.458 &.550  &.538 &.613 &.705 &.751 \\

UNet++ \cite{zhou2019unet++} &.314 &.377 &.447 &\cellcolor{B}{\textbf{\textcolor{white}{.603}}}  &.536 &.616 &\cellcolor{B}{\textbf{\textcolor{white}{.734}}} &.723 \\

ResUNet++ \cite{jha2019resunet}  &.204 &.275 &.464 &.303  &.282 &.368 &.622 &.353 \\

MSEG \cite{huang2021hardnet}   &.309 &.377 &.459 &.525  &.555 &.634 &.720 &\cellcolor{B}{\textbf{\textcolor{white}{.772}}}  \\


LDNet \cite{zhang2022lesion}    &\cellcolor{B}{\textbf{\textcolor{white}{.326}}} &\cellcolor{B}{\textbf{\textcolor{white}{.403}}} &\cellcolor{R}{\textbf{\textcolor{white}{.494}}} &.562  &\cellcolor{B}{\textbf{\textcolor{white}{.604}}} &\cellcolor{B}{\textbf{\textcolor{white}{.675}}} &\cellcolor{R}{\textbf{\textcolor{white}{.770}}} &.767 \\

TGANet \cite{tomar2022tganet} &.253 &.329 &\cellcolor{B}{\textbf{\textcolor{white}{.465}}} & .419  &.374 &.454 &.602 &.505 \\

XBFormer \cite{wang2023xbound} &\cellcolor{R}{\textbf{\textcolor{white}{.360}}} &\cellcolor{R}{\textbf{\textcolor{white}{.421}}} &.451 &\cellcolor{R}{\textbf{\textcolor{white}{.777}}}  &\cellcolor{R}{\textbf{\textcolor{white}{.634}}} &\cellcolor{R}{\textbf{\textcolor{white}{.692}}} &.678 &\cellcolor{R}{\textbf{\textcolor{white}{.943}}} \\

\Xhline{1pt}
 \end{tabular}  }}
\vspace{0.5em}
\caption{Results of the models trained on Kvasir-SEG and tested on multi-centre colonoscopy dataset PolypGen.}
\label{tab:generalization}
\end{table}
\end{center}

\section{Performance Benchmarking}
\label{sec:performance}

This section presents empirical analyses to illustrate essential challenges and progress in the CPS field. Firstly, we evaluate the current state-of-the-art (SOTA) polyp segmentation models and report their performance on five commonly used benchmark datasets (Sec. \ref{subsection:sota}). Secondly, we assess SOTA polyp segmentation models' generalizability on out-of-distribution datasets belonging to different medical centers (Sec. \ref{subsection:ood}). Subsequently, we undertake an attribute-based study to understand better the inherent strengths and weaknesses in current models (Sec. \ref{subsection:attr}).

\subsection{SOTA Performance Comparisons}
\label{subsection:sota}

Table \ref{tab:results} illustrates the performance of 40 cutting-edge deep CPS models across five widely used benchmark datasets, and the methodologies proposed over the last three years are emphasized. These SOTA models were measured by two commonly adopted metrics, i.e., mDice and mIoU. 
To ensure a fair comparison, we follow the same setting as PraNet \cite{fan2020pranet}, which includes 900 and 548 images from ClinicDB and Kvasir-Seg datasets as the train set, and the remaining 64 and 100 images are used as the test set. For clarity, we highlight the best and second performances using red and blue colors. We aspire that our performance benchmarking will contribute to establishing an open and standardized evaluation system in the CPS community.

As shown in Table \ref{tab:results}, RPFA \cite{su2023revisiting} and PPFormer \cite{cai2022using} achieved the Top 2 performance on CVC-ClinicDB, Kvasir-SEG, CVC-ColonDB, ETIS-Larib and EndoScene datasets. In the CVC-ColonDB and On ETIS-Larib datasets, RPFA demonstrated its superior performance, outperforming PPFormer by 1.4\% and 3.1\% in terms of mDice, respectively. Conversely, for the CVC-ClinicDB and EndoScene datasets, PPFormer achieves the best performance by surpassing the RPFA model by 1.5\% and 1.4\%. As anticipated, the overall learning capabilities of these deep learning-based models continue to improvement over time. For instance, the UNet \cite{ronneberger2015u} segmentation model, introduced in 2015, initially achieved only 82.3\% in terms of mDice on the CVC-ClinicDB dataset. Presently, the recently proposed PPFormer achieves an impressive 94.6\%, signaling an annual improvement in segmentation accuracy of nearly 1.4\%.

\subsection{Out-of-Distribution Generalization}
\label{subsection:ood}
Generalization measures how effectively a well-trained model can be applied to out-of-distribution data. A model with strong generalization can make accurate predictions on unseen data, which is essential for deploying deep learning-based poppy segmentation models in real-world clinical scenarios. To assess the SOTA model's generalization ability, we first employ three datasets not encountered during training: ETIS, CVC-ColonDB, and EndoScene. These datasets comprise a combined total of 196, 300, and 912 images, respectively. Unsurprisingly, as shown in Table \ref{tab:results}, the model exhibits considerably lower performance on ETIS, ColonDB, and EndoScene than on the training datasets (CVC-ClinicDB and Kvasir-SEG). For instance, the performance of RPFA \cite{su2023revisiting} on the CVC-ColonDB dataset is approximately 10\% lower than on the CVC-ClinicDB dataset.

Additionally, we explore the generalization ability of these SOTA models on the PolypGen dataset \cite{ali2023multi}, collected from six different centers representing diverse populations. Consequently, validating these SOTA models on PolypGen enhances the comprehensiveness of the study and brings it closer to real-world scenarios. As shown in Table \ref{tab:generalization}, compared to the ETIS, CVC-ColonDB, and EndoScene dataset, these SOTA models exhibit a more pronounced performance degradation on the PolypGen dataset. In particular, XBFormer \cite{wang2023xbound} achieves 87.5\% in terms of mIoU on CVC-ColinicDB while only 36\% on PolypGen-C5. This performance drop is more substantial compared to ETIS, CVC-ColonDB, and EndoScene datasets. Therefore, we can easily conclude that when the distribution of the test dataset is inconsistent with the training data set, the model performance will decrease significantly, and the degree of model performance degradation is proportional to the degree of inconsistency.

\begin{table}[!t]
  \centering
  \begin{tabularx}{0.5\textwidth}{lX}
    \toprule
  \textbf{Attr.}  &  \textbf{Description}   \\
    \midrule
    \textbf{SI}  & \textit{Surgical Instruments.} The endoscopic surgical procedures involve the positioning of instruments, such as snares, forceps, knives, and electrodes.  \\

    \textbf{IB} & \textit{Indefinable Boundaries.} The foreground and background areas around the object have similar color. \\

    \textbf{HO} & \textit{Heterogeneous Object.} Object regions have distinct colors. \\

    \textbf{GH} & \textit{Ghosting.} Object has anomaly RGB-colored boundary due to fast moving or insufficient refresh rate. \\

    \textbf{FM} &  \textit{Fast-motion.}  The average per-frame object motion in a clip, computed as the Euclidean distance of polyp centroids
between consecutive frames, is larger than 20 pixels \\

  \textbf{SO} & \textit{Small Object.} The average ratio between the object size and the image area in a clip is smaller than 0.05. \\

  \textbf{LO} & \textit{Large Object.} The average ratio between the object size and the image area in a clip is larger than 0.15. \\

  \textbf{OC}  & \textit{Occlusion.} Polyp object becomes partially or fully occluded. \\

  \textbf{OV} & \textit{Out-of-view.} Polyp object is partially clipped by the image boundaries. \\

  \textbf{SV} & \textit{Scale-variation.} The average area ratio among any pair of bounding boxes enclosing the target object in a clip is
smaller than 0.5. \\

    \bottomrule
  \end{tabularx}%

  \caption{ Visual attributes and descriptions of SUN-SEG dataset.}

  \label{tab:attribute-based}%
\end{table}%

\subsection{Attribute-based Performance Analysis}
\label{subsection:attr}

\begin{center}
\begin{table*}[t]
\setlength{\abovecaptionskip}{0pt}%
\setlength{\belowcaptionskip}{5pt}%
\large{
\linespread{2}
\renewcommand\arraystretch{1}
\resizebox{0.99\textwidth}{!}{
\begin{tabular}{r||rccccccccc||ccccccccccccc}
\Xhline{1pt}
\multirow{2}{*}{Method}

& \multicolumn{10}{c||}{SUN-SEG Easy  }

& \multicolumn{10}{c}{SUN-SEG Hard }


\\
\cline{2-21}
   &                            SI & IB & HO & GH & FM & SO & LO & OC & OV & SV       & SI & IB & HO & GH & FM & SO & LO & OC & OV & SV \\
\hline\hline

UNet \cite{ronneberger2015u}    & .675 & .548 & .768 & .715 & .633 &.593 &.648 &.670 &.643 &.620 &.618 &.619 &.663 &.676 &.713 &.689 &.633 &.658 &.659 &.658         \\

UNet++ \cite{zhou2019unet++}  &.701 &.542 &.782 &.739 &.647 &.591 &.678 &.683 &.665 &.617 &.654 &.604 &.665 &.696 &.714 &.681 &.660 &.676 &.677 &.678 \\

COSNet  \cite{lu2019see}          &.663 &.531 &.786 &.684 &.610 &.549 &.637 &.648 &.613 &.617 &.641 &.593 &.727 &.668 &.690 &.637 &.694 &.707 &.666 &.625 \\

ACSNet \cite{zhang2020adaptive}  &.789 &.612 &\cellcolor{B}{\textbf{\textcolor{white}{.896}}} &.820 &.704 &.663 &.787 &.770 &.759 &.705 &\cellcolor{B}{\textbf{\textcolor{white}{.770}}} &.681 &\cellcolor{B}{\textbf{\textcolor{white}{.828}}} &\cellcolor{B}{\textbf{\textcolor{white}{.795}}} &\cellcolor{B}{\textbf{\textcolor{white}{.817}}} &.738 &\cellcolor{B}{\textbf{\textcolor{white}{.810}}} &\cellcolor{R}{\textbf{\textcolor{white}{.828}}} &.806 &.759 \\

PraNet \cite{fan2020pranet}  &.745 &.585 &.821 &.772 &.673 &.611 &.722 &.722 &.703 &.653 &.673 &.635 &.725 &.720 &.755 &.691 &.666 &.714 &.708 &.703  \\

SANet \cite{wei2021shallow}  &.724 &.582 &.854 &.760 &.676 &.615 &.703 &.701 &.711 &.680 &.658 &.565 &.738 &.709 &.760 &.692 &.733 &.729 &.727 &.693 \\

MAT \cite{zhou2020matnet} &.772 &\cellcolor{B}{\textbf{\textcolor{white}{.664}}} &.873 &.789 &.706 &\cellcolor{R}{\textbf{\textcolor{white}{.691}}} &.755 &.738 &.746 &.715 &\cellcolor{R}{\textbf{\textcolor{white}{.772}}} &\cellcolor{B}{\textbf{\textcolor{white}{.701}}} &.801 &.776 &.782 &\cellcolor{B}{\textbf{\textcolor{white}{.780}}} &.791 &.795 &.789 &.750  \\

PCSA \cite{gu2020pyramid} &.676 &.563 &.759 &.708 &.628 &.610 &.634 &.662 &.656 &.616 &.656 &.591 &.692 &.683 &.706 &.671 &.612 &.677 &.665 &.663  \\

AMD \cite{liu2021emergence} &.476 &.461 &.471 &.481 &.484 &.466 &.447 &.467 &.442 &.498 &.471 &.468 &.447 &.473 &.468 &.469 &.453 &.487 &.462 &.481  \\

DCF \cite{zhang2021dynamic} &.465 &.485 &.479 &.505 &.541 &.495 &.362 &.484 &.492 &.495 &.441 &.508 &.422 &.498 &.587 &.556 &.351 &.470 &.494 &.540  \\

FSNet \cite{ji2021full} &.719 &.603 &.810 &.752 &.694 &.632 &.686 &.711 &.691 &.665 &.662 &.648 &.743 &.713 &.774 &.723 &.701 &.728 &.728 &.694  \\

PNSNet \cite{ji2021progressively} &.789 &.592 &.871 &.820 &.723 &.619 &.768 &.749 &.751 &.705 &.746 &.631 &.803 &.780 &.778 &.743 &.805 &.790 &.794 &.758 \\

HBNet \cite{puyal2020endoscopic} &\cellcolor{B}{\textbf{\textcolor{white}{.809}}} &.625 &\cellcolor{R}{\textbf{\textcolor{white}{.899}}} &\cellcolor{B}{\textbf{\textcolor{white}{.835}}} &\cellcolor{B}{\textbf{\textcolor{white}{.728}}} &.667 &\cellcolor{R}{\textbf{\textcolor{white}{.820}}} &\cellcolor{R}{\textbf{\textcolor{white}{.783}}} &\cellcolor{B}{\textbf{\textcolor{white}{.778}}} &\cellcolor{B}{\textbf{\textcolor{white}{.719}}} &.768 &.662 &\cellcolor{R}{\textbf{\textcolor{white}{.865}}} &.784 &.797 &.737 &\cellcolor{R}{\textbf{\textcolor{white}{.853}}} &\cellcolor{B}{\textbf{\textcolor{white}{.827}}} &\cellcolor{R}{\textbf{\textcolor{white}{.808}}} &\cellcolor{B}{\textbf{\textcolor{white}{.765}}}  \\

PNS+ \cite{ji2022video} &\cellcolor{R}{\textbf{\textcolor{white}{.819}}} &\cellcolor{R}{\textbf{\textcolor{white}{.667}}}&.883 &\cellcolor{R}{\textbf{\textcolor{white}{.844}}} &\cellcolor{R}{\textbf{\textcolor{white}{.738}}} &\cellcolor{B}{\textbf{\textcolor{white}{.690}}} &\cellcolor{B}{\textbf{\textcolor{white}{.796}}} &\cellcolor{B}{\textbf{\textcolor{white}{.782}}} &\cellcolor{R}{\textbf{\textcolor{white}{.798}}} &\cellcolor{R}{\textbf{\textcolor{white}{.734}}} &\cellcolor{B}{\textbf{\textcolor{white}{.770}}} &\cellcolor{R}{\textbf{\textcolor{white}{.703}}} &.817 &\cellcolor{R}{\textbf{\textcolor{white}{.801}}} &\cellcolor{R}{\textbf{\textcolor{white}{.823}}} & \cellcolor{R}{\textbf{\textcolor{white}{.793}}} &.792 &.808 &\cellcolor{B}{\textbf{\textcolor{white}{.807}}} &\cellcolor{R}{\textbf{\textcolor{white}{.795}}} \\

\Xhline{1pt}
\end{tabular}  }}
\vspace{0.5em}
\caption{Visual attributes-based performance comparisons on SUN-SEG-Easy/Hard using structure measure.}
\label{tab:attribution-analysis}
\end{table*}
\end{center}

\begin{figure*}[t]
\centering
\resizebox{1\textwidth}{!}{
\includegraphics[width=\textwidth]{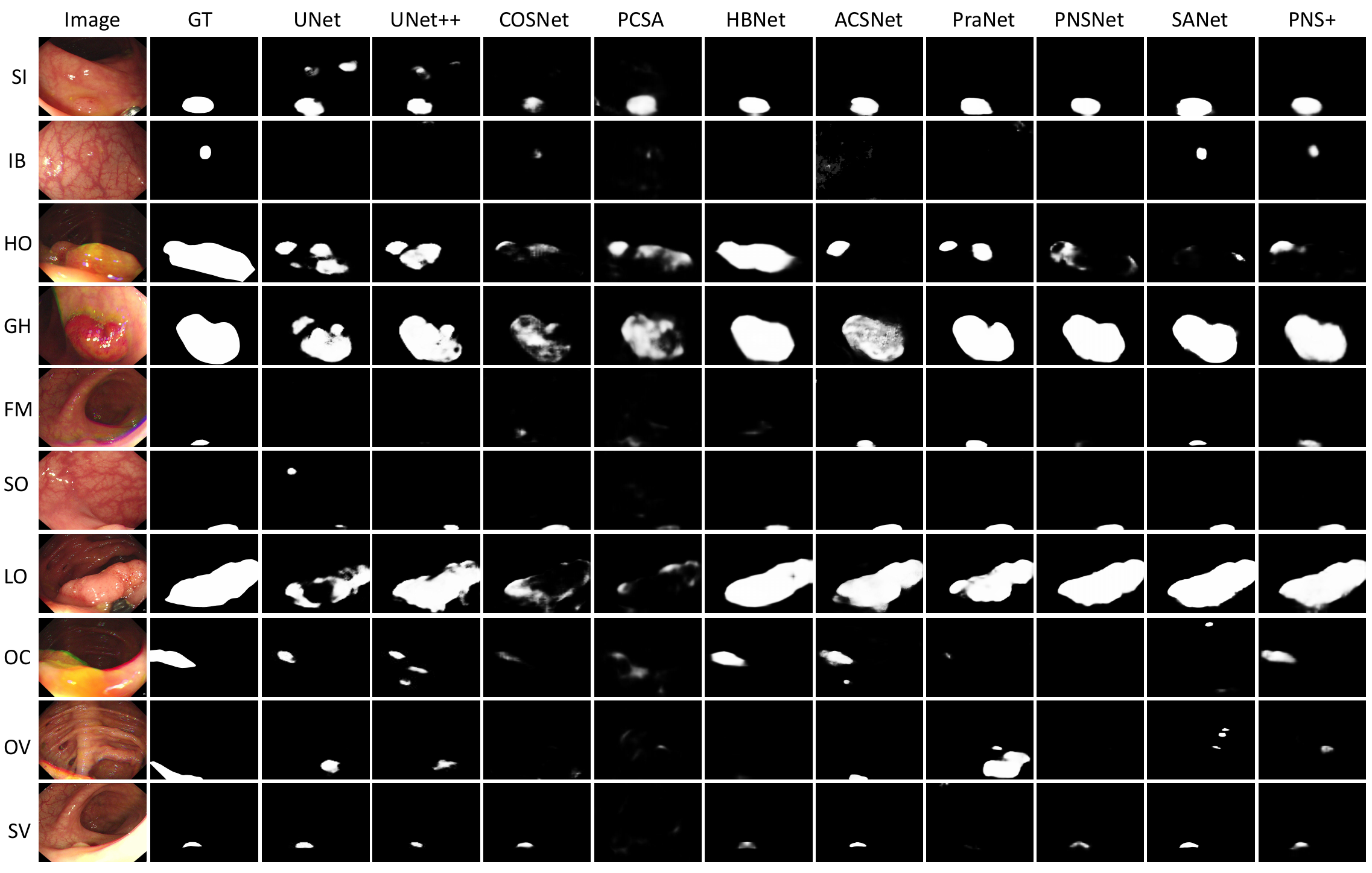}
}\vspace{-0.5em}
\caption{Qualitative visualization of the selected representative deep CPS models on SUN-SEG dataset.} 
\label{fig:visual}
\end{figure*}

While the community has observed significant progress in deep CPS models, there remains a lack of clarity regarding the specific conditions under which these models perform well. Given the multitude of factors influencing CPS algorithm performance, including scene category and occlusion, it becomes imperative to assess their effectiveness across diverse scenarios. This evaluation is essential for uncovering deep CPS models' nuanced strengths and weaknesses, identifying unresolved challenges, and pinpointing prospective research avenues to develop more robust algorithms. This paper employs the recently proposed SUN-SEG dataset \cite{ji2022video}, containing 158 690 frames selected from the SUNdatabase \cite{misawa2021development}. Ji \textit{et al.} \cite{ji2022video} divided the SUN-SEG dataset into SUN-SEG-Easy with 119 clips (17, 070 frames) and SUN-SEG-Hard with 54 clips (12, 522 frames) according to difficulty levels in each pathological category, and further provide ten visual attributes according to the visual characteristics of images, as detailed in Table \ref{tab:attribute-based}.

In Table \ref{tab:attribution-analysis}, we report the performance of specific attribute-based subsets on the SUN-SEG dataset. For clarity, we highlight the best and second performances using red and blue colors. Besides, as shown in Fig. \ref{fig:visual}, we also present visual results on these attribute-based scenarios. Below are some important observations drawn from these results.

\noindent \textbf{1) Not winner takes all.} Among the compared methods, PNS+ \cite{ji2022video} stands out as the best from an overall perspective; however, it falls short of achieving optimal performance in every attribute. Similarly, while the ACSNet \cite{zhang2020adaptive} does not secure a spot in the top three rankings, it excels in achieving the best performance for specific attributes.

\noindent \textbf{2) Easy and hard scenes.} Unexpectedly, methods tested on the SUN-SEG easy dataset achieved better performance than those tested on the SUN-SEG hard dataset regarding SI, HO, and GH attributes. Typically, the model's performance on the SUN-SEG easy dataset should surpass that on the SUN-SEG hard dataset, indicating potential errors in the  SUN-SEG dataset partitioning.

\noindent \textbf{3) Easy and hard attribution.} For both SUN-SEG easy and SUN-SEG hard datasets, attribute HO is identified as the least challenging, with HBNet \cite{puyal2020endoscopic} achieving a performance score of 0.899. Conversely, attribute IB is considered the most challenging, and the PNS+'s \cite{ji2022video} peak performance on attribute IB only reaches 0.667.


\section{Challenges and Future Directions}
\label{sec:challenges}

Undoubtedly, deep learning has substantially advanced polyp segmentation; however, there are numerous challenges that need to be addressed. In the following sections, we will explore promising research directions that we believe will help in further advancing polyp segmentation.

\subsection{Interpretable Deep CPS Model}
The developments of deep neural networks have revolutionized the fields of artificial intelligence, and have achieved promising performance in polyp segmentation tasks.
Nevertheless, the majority of deep models are designed without prioritizing interpretability, making them as black-box systems. The absence of a clear understanding of the underlying mechanisms behind predictions raises concerns about the trustworthiness of these deep models. This lack of transparency impedes their application in real-world clinical scenarios. For instance, why does the model identify the image as a polyp? What are the criteria for this decision? To ensure the secure and trustworthy deployment of deep models, it becomes imperative to deliver not only accurate predictions but also human-intelligible explanations, particularly for users in medical diagnosis. These factors mentioned above underscore the necessity for the development of novel techniques to interpret deep neural networks.

\subsection{Federated Learning for Data Privacy}

Current deep learning-based models are data-hungry, which means that the greater the volume of training data, the higher the model's performance. However, the reality is that medical data is often scattered across different hospitals, safeguarded by privacy restrictions. For example, data from different hospitals are isolated and become "data islands." Given the constraints in size and distribution within each data island, a single hospital may struggle to train a high-performance model. In an ideal scenario, hospitals can benefit more if we can collaboratively train a deep learning model on the union of their data. However, the challenge is that we cannot straightforwardly share data among hospitals due to data privacy and diverse policies. The recently emerged federated learning \cite{kairouz2021advances,hong2021communication,10040221,10274722}can handle these challenges, which can collaboratively train machine learning models without collecting their local data, making it particularly suitable for scenarios where data cannot be easily centralized due to privacy concerns.

\subsection{Domain Adaptation for Domain Shift}

The existing deep learning-based CPS model typically assumes that the training dataset (source/reference domain) and the test dataset (target domain) share the same data distribution. Unfortunately, this assumption is overly restrictive and may not be true in real-world scenarios. Owing to various factors such as illumination and image quality, a distribution shift commonly occurs between the training and testing datasets that can degrade the performance drastically. As demonstrated in Table \ref{tab:results} and Table \ref{tab:generalization}, when the distribution of the test dataset is inconsistent with the training data set, these SOTA model's performance decreased significantly. Thus, handling domain shift is crucial to effectively applying deep learning methods to medical image analysis. As a promising solution to tackle the distribution shift among medical image datasets, domain adaptation \cite{kouw2019review,dong2021and,oza2023unsupervised} has attracted increasing attention in many tasks, aiming to minimize the distribution gap among different but related domains. 

\subsection{Defending Against Adversarial Attack}
Recent research revealed that deep models are susceptible to adversarial attacks, which may be invisible to the human eye but can lead the model to misclassify the output. Szegedy \textit{et al.} \cite{szegedy2014intriguing} was the first to demonstrate that high-performing deep neural networks can also fall prey to adversarial attacks. Su \textit{et al.} \cite{su2019one} claimed successful fooling of three different network models on the tested images by changing only one pixel per image. Additionally, they observed that the average confidence of the networks in assigning incorrect labels was 97.47\%. Such an adversarial attack poses a significant challenge to deploying deep CPS models in real-world clinical applications. In particular, the model's accurate and reliable diagnoses are extremely important in healthcare because misdiagnosis may cause severe consequences, including patient mortality. Hence, it is extremely urgent to improve the robustness of the deep CPS model against various adversaries.

\subsection{Weakly/Un-supervised Learning}

As shown in Table \ref{tab:methods1} and Table \ref{tab:methods2}, existing deep CPS models are commonly trained in a fully-supervised manner, relying on numerous meticulously annotated pixel-level groundtruths. However, constructing such a well-annotated pixel-level dataset is resource-intensive and time-consuming. Recently, there have been some approaches to training a CPS model with weak supervision, such as bounding-box \cite{wei2023weakpolyp} and scribble level \cite{wang2023s2me} annotations, but performance disparity remains compared to a fully-supervised model. Another noteworthy direction is self-supervised learning, which has gained considerable attention across various tasks. Misra \textit{et al.} \cite{misra2020self} have proved that a self-supervised model can capture intricate image details, facilitating the training of segmentation models. With the multitude of algorithmic breakthroughs witnessed in recent years, we anticipate a surge of innovation in this promising direction.

\subsection{Lightweight Model for Real-world Application}

Advanced deep CPS models are intricately designed to enhance learning capacity and the model's performance. However, there is a growing requirement for more innovative and lightweight architectures to meet the demands of mobile and embedded devices. Dollar \textit{et al.} \cite{Dollar_2021_CVPR} demonstrated that simply scaling the model capacity significantly incurs performance degradation in terms of accuracy and generalization. To facilitate the practical implementation of deep CPS models in clinical settings, we recommend developing a lightweight model to maintain a good balance between performance and efficiency. Another recommended direction is employing model compression \cite{choudhary2020comprehensive} or knowledge distillation \cite{li2020few} to condense these heavy and high-performance models, which can ensure minimal performance degradation and obtain a lightweight model.

\subsection{Connecting CPS with Anomaly Detection}
Anomaly detection \cite{xia2022gan}, also known as outlier detection, aims to identify patterns or instances that deviate significantly from the norm or expected behavior within a dataset. The goal is to identify data points that are considered rare, unusual, or suspicious compared to the majority of the data. In the polyp dataset, a significant portion of images does not contain polyps, with only a subset containing polyps. Thus, polyps in an image can be considered typical instances of anomaly detection, which means that we can segment polyps by using unsupervised anomaly localization techniques without large-scale pixel-wise annotations. Recently, some researchers have trained the polyp segmentation model with the help of unsupervised anomaly detection, i.e., CCD \cite{tian2021constrained} and PMSACL \cite{tian2023self}. We believe there is still much room for improvement in this direction, and the prospects for unsupervised polyp localization are promising.

\subsection{Combining CPS with Large Segmentation Models}

Very recently, the segmentation anything model (SAM) \cite{kirillov2023segment} has gained massive attention due to its impressive performance in many segmentation tasks, which has been trained on the largest segmentation dataset with more than 1 billion image-mask pairs, surpassing existing segmentation datasets by a factor of 400. Finetuning the SAM for downstream tasks, such as camouflaged object detection \cite{tang2023can}, skin cancer segmentation \cite{hu2023skinsam} and image style transfer \cite{yu2023inpaint}, has become a hot research area. Zhou \textit{et al.} \cite{zhou2023can} show that directly applying SAM to the polyp segmentation cannot achieve satisfactory performance for these unseen medical images. Thus, one promising direction is to finetune the SAM model using training polyp datasets.

\subsection{Combining CPS with Large Language Models}
Combining large language models (LLMs)  and computer vision has recently become a hot research area, driving significant advancements in the many tasks \cite{menon2022visual,Parisot_2023_CVPR}. The large language models,  initially created to understand human language, are gradually expanding to encompass visual tasks and begin to combine text data with visual data. This fusion of LLMs and CV leads to an era where AI systems can see the world, understand it, and communicate with it just like humans do. In this way, the deep model detects lesions and explains them clearly, helping doctors understand and trust the prediction results. On the other hand, combining LLMs and polyp segmentation could lead to better diagnosis and treatment by joining visual information with a wide range of medical knowledge. In summary, this is a promising direction and has the potential to break the current learning paradigm.

\section{Conlusion}
\label{sec:conlusion}
This paper presents a comprehensive review of deep learning-based CPS models. Firstly, we provide innovative taxonomy for categorizing deep CPS models from perspectives of network architecture, levels of supervision, and learning paradigms. Subsequently, we delve into contemporary literature on popular CPS datasets and evaluation metrics and conduct a thorough performance benchmarking of major CPS methods.
In particular, we reveal the strengths and weaknesses of CPS datasets by comparing the number of datasets available, types of annotations, image resolutions, polyp sizes, contrast values, and polyp location. Furthermore, we evaluate the model's generalization performance on out-of-distribution datasets and its attribute-based performance on SUN-SEG datasets, providing a nuanced understanding of the strengths and weaknesses of deep CPS models.
Finally, we look deeper into the challenges of current deep learning-based CPS models, providing insightful discussions and several potentially promising directions. We hope our survey will help researchers gain a deeper understanding of the developmental history of CPS and inspire new works to advance this field.


\if CLASSOPTIONcaptionsoff
\newpage
\fi

\bibliographystyle{IEEEtran}
\bibliography{aaai22}

\vfill

\end{document}